 \documentclass[times,twocolumn,final]{elsarticle}
\usepackage{jasr}
\usepackage{framed,multirow}
\usepackage{graphicx}

\usepackage{amssymb}
\usepackage{latexsym}

\usepackage{url}
\usepackage{xcolor}
\definecolor{newcolor}{rgb}{.8,.349,.1}

\usepackage[citebordercolor=white]{hyperref}

\journal{Advances in Space Research}

\begin{document}

\verso{Sandeep Nagar \textit{etal}}

\begin{frontmatter}

\title{Remote sensing framework for geological mapping via stacked autoencoders and clustering}

\author[1]{Sandeep  Nagar}
\author[2]{Ehsan  Farahbakhsh\fnref{fn1}}
\fntext[fn1]{This author contributed equally to the paper as the first author.}
\author[3]{Joseph  Awange}
%% Third author's email
% \ead{author3@author.com}
\author[4]{Rohitash  Chandra\corref{cor1}}
\cortext[cor1]{Corresponding author: 
  Email: rohitash.chandra@unsw.edu.au}

\affiliation[1]{Machine Learning Lab, International Institute of Information Technology, Hyderabad, India}
\affiliation[2]{EarthByte Group, School of Geosciences, The University of Sydney, Sydney, Australia}
\affiliation[3]{School of Earth and Planetary Sciences, Curtin University, Perth, Australia}
\affiliation[4]{Transitional Artificial Intelligence Research Group, School of Mathematics and Statistics, University of New South Wales, Sydney, Australia}

% \received{1 May 2013}
% \finalform{10 May 2013}
% \accepted{13 May 2013}
% \availableonline{15 May 2013}
% \communicated{S. Sarkar}

\begin{abstract}
Supervised machine learning methods for geological mapping via remote sensing face limitations due to the
scarcity of accurately labelled training data that can be addressed by unsupervised learning, such as
dimensionality reduction and clustering. Dimensionality reduction methods have the potential to play a
crucial role in improving the accuracy of geological maps. Although conventional dimensionality reduction
methods may struggle with nonlinear data, unsupervised deep learning models such as autoencoders can
model non-linear relationships. Stacked autoencoders feature multiple interconnected layers to capture
hierarchical data representations useful for remote sensing data.  We present an unsupervised
machine learning-based framework for processing remote sensing data using stacked autoencoders for
dimensionality reduction and k-means clustering for mapping geological units. We use Landsat 8, ASTER,
and Sentinel-2 datasets to evaluate the framework for geological mapping of the Mutawintji region in
Western New South Wales, Australia. We also compare stacked autoencoders with principal component
analysis (PCA) and canonical autoencoders. Our results reveal that the framework produces accurate and
interpretable geological maps, efficiently discriminating rock units. The results reveal that the combination
of stacked autoencoders with Sentinel-2 data yields the best performance accuracy when compared to other
combinations. We find that stacked autoencoders enable better extraction of complex and hierarchical
representations of the input data when compared to canonical autoencoders and PCA. We also find that the generated maps align with prior geological knowledge of the study area while providing novel insights into geological structures.

\end{abstract}


\end{frontmatter}
 
%% main text
\section{Introduction}

Geological mapping is crucial for various purposes, such as assessing the mineralization potential of a region and creating prospectivity maps \citep{bachri2019machine,wang2021lithological,shirmard2022review}. Remote sensing provides an efficient alternative to traditional fieldwork for geological mapping, which is often costly and sometimes impractical due to harsh topography or political constraints \citep{yu2012towards}. Optical remote sensing images, captured by scanners on platforms such as satellites, cover multiple spectral bands ranging from the visible to infrared regions of the electromagnetic spectrum \citep{clark2003imaging}. The detailed information in multispectral images aids geological analysis by identifying and mapping rocks and minerals based on their specific spectral absorption properties \citep{chen2010integrating,weilin2016application,lu2021lithology}. This spectral data forms the foundation for spectrum-based approaches in classifying and mapping image pixels. The spectral and spatial resolution of remote sensing data, especially from satellites, allows for the identification of rock units over extensive areas \citep{pour2018mapping}. However, geological mapping via remote sensing, particularly using optical data, faces different challenges due to the influence of land cover, which can obscure rock outcrops, introduce spectral mixing, and vary seasonally, complicating the interpretation of spectral signatures. Geological mapping focuses on identifying and characterizing rock types, geological formations, and structures to understand Earth's history and resources, while land use/land cover mapping classifies and monitors human activities and natural features on the Earth's surface for urban planning, agriculture, and environmental management \citep{bouslihim2022comparing,tabassum2023exploring,akanbi2024integrating,amare2024impacts,masoudi2024assessment}. Land cover types such as vegetation and urban structures can mask or mimic rock characteristics, and shadows can mislead classifications \citep{hashim2013automatic,bentahar2020fracture,shebl2021reappraisal,shirmard2022review}. However, integrating optical data with other remote sensing data types and employing advanced data analytics tools enhances accuracy, making optical data a valuable tool for geological mapping despite the challenges posed by land cover \citep{galdames2019rock,ran2019rock}.

Common geological mapping techniques based on remote sensing data processing involve comparing absorption features to reference spectra or training samples \citep{chen2010integrating}. However, collecting sufficient pixels for reference spectra or training samples is challenging. Geological processes influence the spectral variability in rocks, which depends on their chemical and mineral composition, grain size, texture, and structure \citep{sgavetti2006reflectance}. This spectral variability significantly impacts geological mapping using remote sensing data. Insufficient training samples with highly correlated spectral bands often lead to challenges in discriminating rock units \citep{bruzzone2014review}. Therefore, extracting useful information from images and removing redundant information is essential to improve geological discrimination \citep{sun2019hyperspectral}. Machine learning models, particularly deep learning models, have proven to be powerful tools for extracting valuable information from remote sensing data and mapping geological anomalies \citep{zhao2023recognition,dou2024large}. Such models utilize dimensionality reduction, clustering, and classification approaches to automatically analyze and interpret complex datasets \citep{bedini2009mapping,carneiro2012semiautomated,sahoo2017pattern,awange2020hybrid,dou2024remote}. In remote sensing, machine learning models have found widespread applications \citep{zuo2019deep,dou2024time}, and their adoption in mineral exploration is gaining significant traction \citep{shirmard2022comparative}. These models have been integrated with conventional image processing techniques and geological surveys, significantly enhancing remote sensing for geological mapping and mineral prospectivity mapping \citep{shirmard2022review,guo2023gis,hajaj2024review}.

Unsupervised machine learning techniques, such as dimensionality reduction and transformation methods, have shown remarkable efficiency in distinguishing between geological units, making them invaluable for identification and mapping purposes \citep{behnia2012remote}. Data compression and transformation techniques such as principal component analysis (PCA) \citep{wold1987principal}, independent component analysis (ICA) \citep{comon1994independent,forootan2012independent}, and minimum noise fraction (MNF) \citep{nielsen2010kernel} have the ability to suppress irradiance that dominates the bands of remote sensing data \citep{gao2017optimized}, thereby enhancing the spectral reflectance of geological features \citep{richards1999remote}. Supervised learning models can partially automate the extraction of features specifically related to the labelled data and address some of the challenges of semi-manual analysis. However, supervised learning requires ground truth data and expert input, which are costly and can introduce bias \citep{gewali2018machine}. The availability of fully labelled training data for geological features is limited, compounding the challenges faced by supervised learning. Supervised learning models have been applied to multivariate datasets, such as multispectral remote sensing data, to extract specific spectral responses from different rock units \citep{asadzadeh2016review,nalepa2020unsupervised}; however, the large spectral variability and limited availability of training data make classification a challenging task in the geological analysis of remotely sensed data.

Autoencoders are a class of machine learning models primarily used for unsupervised learning tasks such as dimensionality reduction and feature learning \citep{kramer1992autoassociative,kingma2019introduction,li2023comprehensive}. They consist of two main components: an encoder and a decoder. Autoencoders provide a lower-dimensional (reduced) representation of the data \citep{wang2016auto,kingma2019introduction} using a latent vector that enables the data to be represented with fewer features. The latent vector has been used in remote sensing data processing to extract features or structures, such as geological units \citep{protopapadakis2021stacked}, and studies have shown that the features in the latent vector correspond to different types of minerals \citep{calvin2018band,gao2021generalized}. Latent vectors play a crucial role in maintaining feature independence, which is essential for preventing the mixing of different mineral types \citep{protopapadakis2021stacked}. This separation ensures that each mineral type or rock unit is accurately represented and distinguished, enhancing the clarity and reliability of the analysis.

Stacking is an ensemble learning approach \citep{sagi2018ensemble} that has been shown to enhance the performance of autoencoders. Stacked autoencoders \citep{vincent2010stacked} have proven useful in various applications, such as feature extraction for multi-class change detection in hyperspectral images \citep{lopez2018stacked} and the classification of multispectral and hyperspectral images \citep{ozdemir2014hyperspectral,lv2017remote}. They provide a powerful framework for learning deep data representations in an unsupervised manner, which is particularly beneficial for machine learning tasks where labelled data is scarce or expensive to obtain \citep{vincent2010stacked}. In a recent study, \cite{protopapadakis2021stacked} addressed noise in input signals due to dimensionality redundancy without losing important features using a stacked autoencoder. Additionally, stacked autoencoders can help in understanding the nonlinearity between spectral bands and distinguishing complex features, such as geological units \citep{dai2023optimization}.

The extracted latent vectors from dimensionality reduction or transformation techniques can be used as inputs for clustering methods, such as $k$-means clustering, to group pixels and map geological units \citep{davies1979cluster,gao2021generalized}. Clustering methods are unsupervised learning techniques \citep{davies1979cluster,omran2007overview} that organize a dataset into groups based on the similarity of samples (instances) using various distance measures \citep{xie2016unsupervised,yadav2019study}. These methods have been widely used in remote sensing applications alongside dimensionality reduction techniques \citep{rodarmel2002principal}. Examples include pixel clustering \citep{bandyopadhyay2007multiobjective}, fuzzy clustering for change detection \citep{ghosh2011fuzzy}, image segmentation \citep{fan2009single}, mean-shift clustering of multispectral imagery \citep{bo2009mean}, and hyperspectral image subspace clustering involving dimension reduction, subspace identification, and clustering \citep{zhang2016spectral}. However, clustering methods face challenges due to the curse of dimensionality, which complicates their use in remote sensing data processing.

In large datasets, such as multispectral remote sensing data covering vast areas, the combined power of autoencoders and clustering can address the limitations and challenges of conventional methods. A major challenge in remote sensing-based geological mapping arises from the intricate and diverse nature of geological features, coupled with the often remote and inaccessible locations of target areas \citep{pal2020optimized,dos2021deep,shirmard2022review}. Although conventional dimensionality reduction methods may struggle with nonlinear data, autoencoders have been effective in modelling non-linear relationships. Stacked autoencoders, featuring multiple interconnected layers that capture hierarchical data representation, can be useful for remote sensing data processing. Leveraging autoencoders with clustering methods has the potential to provide accurate geological maps.

In this study, we present an unsupervised machine learning framework that combines stacked autoencoders for dimensionality reduction with $k$-means clustering to map geological units, aiding in the identification of potential mineralized areas. The role of the stacked autoencoders is to compress data with a wide range of spectral and spatial features, enhancing both the accuracy and efficiency of geological mapping. We use $k$-means clustering in our framework to generate clustered maps from the reduced dimensions. We evaluate our framework across multiple multispectral remote sensing datasets (Landsat 8, ASTER, and Sentinel-2) to map geological units in the Mutawintji region of Western New South Wales (NSW), Australia. We also compare our results with PCA and canonical autoencoders and provide open-source code to further extend the study. 

\section{Materials and methods}
\subsection{Geological setting}

\begin{figure*}
    \centering
    \includegraphics[width=0.6\textwidth]{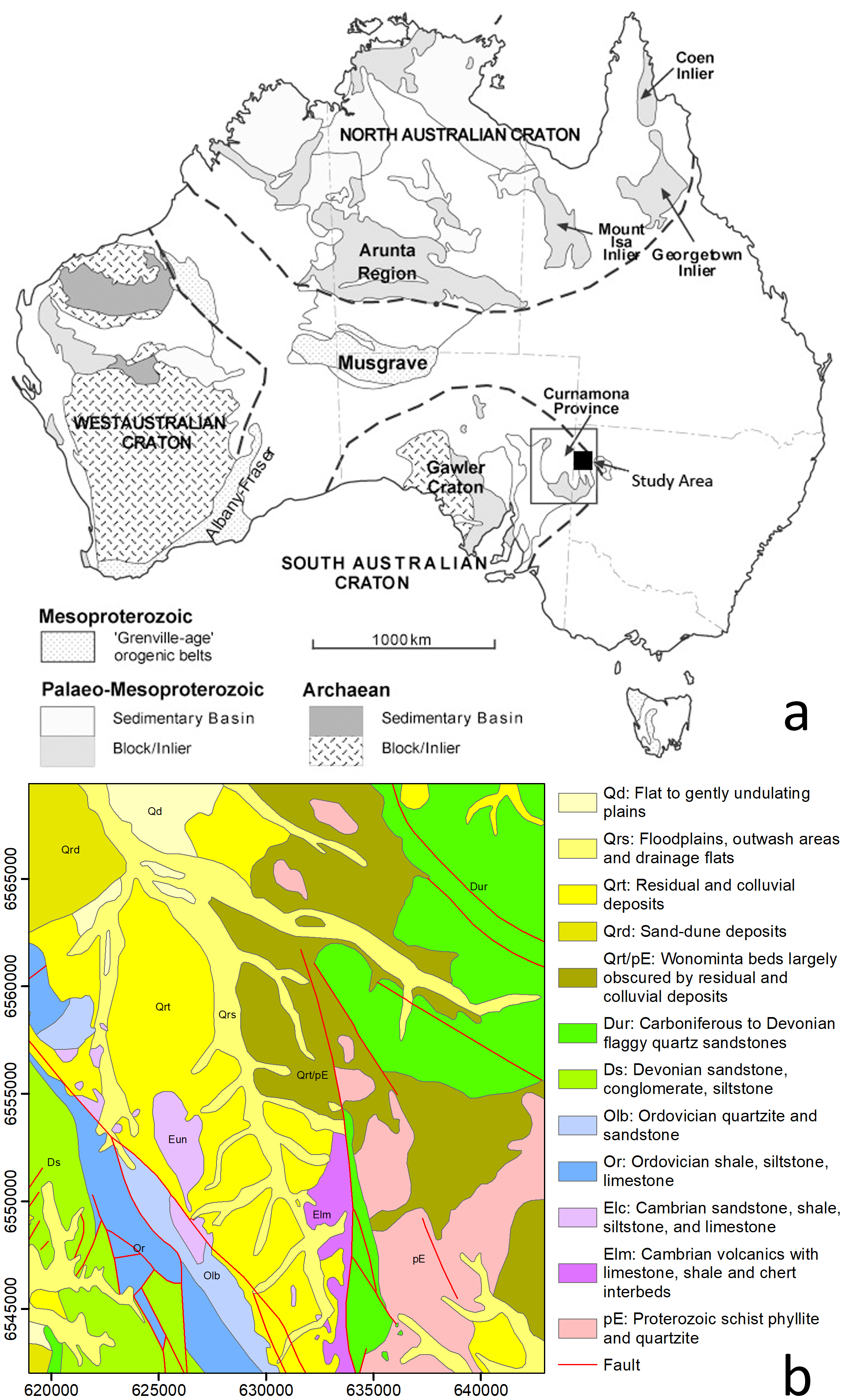}
    \caption{a) The Curnamona Province and other Proterozoic terrains in Australia \citep{barovich2008tectonic}; the study area has been shown using a black square. b) Simplified geological map of the study area.}
    \label{fig_1}
\end{figure*}

The Mutawintji region is located in the far west of NSW and the semi-arid zone of the state. It is approximately 1,150 kilometres (km) west of Sydney and covers an area of approximately 700 $km^2$ within the Curnamona Province, a geological province that covers a large area of southeastern Australia. As shown in Fig. \ref{fig_1}(a), the study area is situated on the eastern margin of the Curnamona Province, characterized by a thick sequence of sedimentary rocks deposited in a shallow marine environment during the Cambrian and Ordovician periods \citep{hewson2005seamless}. The geological setting of the study area is dominated by sedimentary rocks of the Cambrian and Ordovician periods, which are around 500 to 480 million years old. However, Quaternary residual and colluvial deposits cover a significant part of the sedimentary rocks \citep{young2009ordovician}. The sedimentary rocks comprise various rock types, including sandstone, shale, siltstone, and limestone. These rocks were formed by the accumulation of sediments in an ancient sea that covered much of the region during the Cambrian and Ordovician periods. The sediment was later buried and compressed, eventually forming the sedimentary rocks that can be observed on the surface. In addition to the sedimentary rocks, the geology of the study area also includes a range of other rock types, including volcanic rocks and granites. Major faults in the study area strike North--South or North West--South East, which separate Ordovician quartzite and sandstone units from shale, siltstone, and limestone in the southwest. Fig. \ref{fig_1}(b) presents the geology of the study area, characterized by a complex and diverse range of rock types that reflect the region's long and varied geological history and make it an interesting area for our study.

\subsection{Remote sensing data and pre-processing}

Our study utilizes three types of multispectral remote sensing data, each with its unique capabilities and resolutions. Landsat 8 (launched in 2013) is equipped with two sensors—the operational land imager (OLI) and the thermal infrared sensor (TIRS). It provides images in 11 different spectral bands, with resolutions ranging from 15 meters (m) for the panchromatic band to 30 m for the visible and near-infrared (VNIR) and shortwave infrared (SWIR) bands. The thermal bands numbered 10 and 11, have a resolution of 100 m \citep{zhang2016integrating}. Geological remote sensing for mapping has significantly improved with the launch of the ASTER sensor on the Terra platform in 1999. ASTER's VNIR bands have a spatial resolution of 15 m, six SWIR bands with a resolution of 30 m, and five thermal infrared bands with a resolution of 90 m \citep{rowan2003lithologic}. Sentinel-2A and Sentinel-2B are twin satellites in sun-synchronous orbit, phased 180 degrees apart. Their onboard multispectral instrument captures data in 13 spectral bands, ranging from VNIR to SWIR, with spatial resolutions varying from 10 to 60 m \citep{drusch2012sentinel2}. The use of these diverse data types will enable us to comprehensively map the geological units in the study area.

In this study, we focus on spectral bands that are particularly important in geological remote sensing due to their characteristic behaviours, such as high absorption or reflectance in different geological units, which allow for the generation of meaningful geological maps. Accordingly, we select seven bands from OLI (bands 1--7), nine bands from ASTER (bands 1--9), and ten bands from Sentinel-2 (bands 2--8, 8a, 11, and 12). We obtained a cloud-free Landsat 8 scene of the study area from the US Geological Survey Earth Resources Observation and Science (USGS EROS) center\footnote{\url{https://earthexplorer.usgs.gov} (accessed on 31 January 2022)}. The image, captured on 5 October 2021, is a level-1T product that has been terrain-corrected. The acquired ASTER image of the study area was captured on 10 August 2001; this cloud-free level-1-precision terrain-corrected product (ASTER\_L1T) was also obtained from the USGS EROS centre. Additionally, we downloaded a cloud-free Sentinel-2A scene of the study area, captured on 19 March 2022, from the European Space Agency via the Copernicus Open Access Hub\footnote{\url{https://scihub.copernicus.eu/} (accessed on 28 March 2022)}. This Sentinel-2 image is a level-1C product that has undergone radiometric and geometric corrections and orthorectification, resulting in top-of-atmosphere reflectance values.

\begin{figure*}
    \centering
    \includegraphics[width=\textwidth]{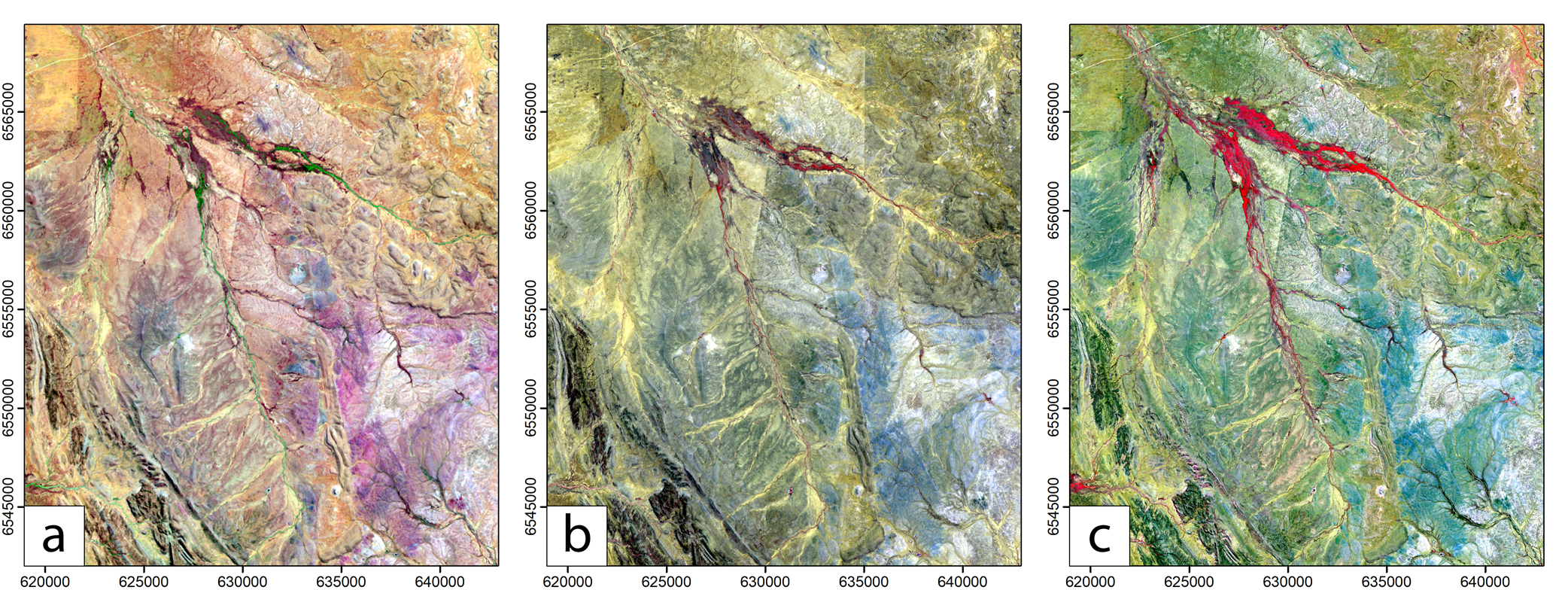}
    \caption{False colour composite images generated using a) Landsat 8 (RGB 753), b) ASTER (RGB 321), and c) Sentinel-2 (RGB 843) data.}
    \label{fig_2}
\end{figure*}

The remote sensing datasets used in this study are pre-georeferenced to the Universal Transverse Mercator (UTM) zone 54 South, eliminating the need for geometric correction. The Landsat 8 OLI and ASTER data have been radiometrically corrected, and the reflectance data are used as inputs to the workflow. To match the spatial resolution of the VNIR bands (15 m), the SWIR bands in the ASTER data are resampled using the nearest neighbour technique \citep{taunk2019brief}. After resampling, a data layer is created by stacking the VNIR and SWIR bands for further processing. The Sentinel-2 image includes atmospheric correction. Using the nearest neighbour technique, we stack the Sentinel-2 VNIR and SWIR bands to create a 10-band dataset with 10 m spatial resolution. Before processing, all images are resized to fit the target area size. Figure \ref{fig_2} shows false colour composite images created by stacking images from the Landsat 8, ASTER, and Sentinel-2 datasets.

\subsection{Stacked autoencoders}

\begin{figure}
    \centering
    \includegraphics[width=0.5\textwidth]{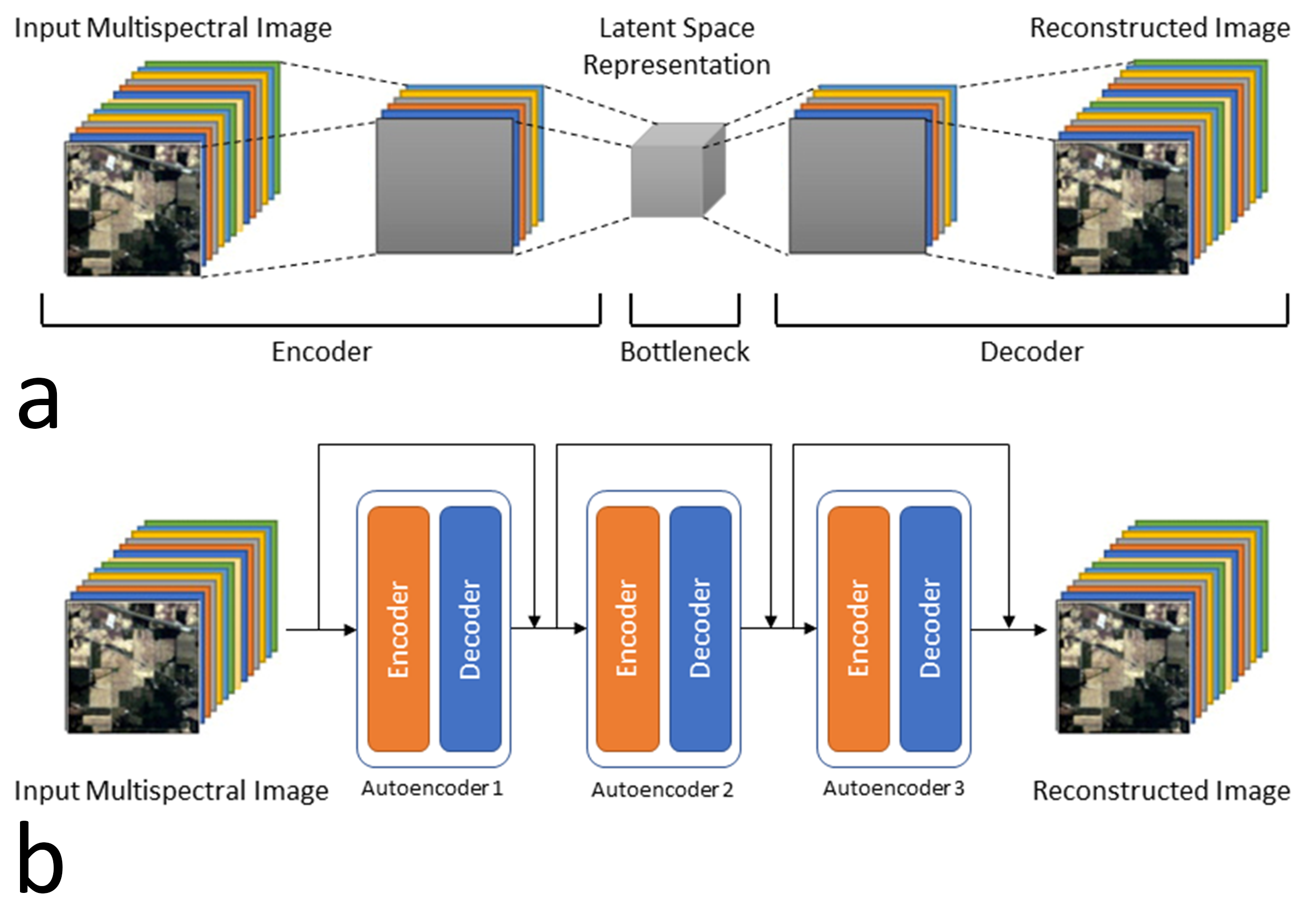}
    \caption{a) The architecture of a canonical autoencoder consists of an encoder and a decoder. The encoder takes the multispectral image as input ($x$) and reduces the dimension to the latent vector ($z$), where dim($x$) $>=$ dim($z$). The decoder reconstructs the image from the latent vector ($Z$). b) A stacked autoencoder with three encoders and decoders for each stacking level. Each stack level's encoder and decoder architecture is the same as the canonical autoencoder, with a number of hidden layers for each.}
    \label{fig_3}
\end{figure}

The autoencoder is a dimensionality reduction technique used to uncover a lower-dimensional manifold, also known as the latent space (vector) of intrinsic dimensionality, while preserving the essential information present in the original data. Fig. \ref{fig_3}(a) depicts a canonical autoencoder consisting of an encoder, which reduces the dimensionality of the input data, and a decoder, which reconstructs the original data from the encoded representation. Unlike conventional dimensionality reduction methods such as PCA, which find linear combinations of the original features \citep{mackiewicz1993principal,abdi2010principal}, autoencoders can learn more abstract and higher-level features of the data. This capability is particularly useful when the data has complex patterns and structures. Although PCA is relatively simpler to implement and use for dimensionality reduction, training autoencoders can be more complex and require more training time and computational resources \citep{lv2017remote}.

\begin{figure}
    \centering
    \includegraphics[width=0.5\textwidth]{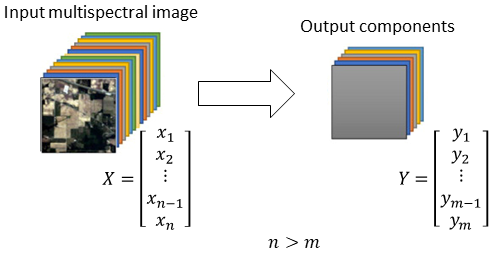}
    \caption{The visualization of the dimensionality reduction for a multispectral dataset. The number of input spectral bands $n$ is reduced to $m$ in the output dataset. Each coloured layer in the input image and the output represents a spectral band and a component, respectively.}
    \label{fig_4}
\end{figure}

The stacked autoencoder architecture comprises multiple layers of autoencoders, each trained independently as an individual autoencoder \citep{xu2019review}, as shown in Fig. \ref{fig_3}(b). The output from one layer serves as the input for the next, enabling the network to learn hierarchical representations of the data. Fig. \ref{fig_3}(b) presents a layered autoencoder with three encoders and decoders stacked sequentially \citep{dai2023optimization}. A typical stacking approach involves at least two layers: the first layer contains several models (any machine learning model), and the second layer combines the predictions using a simpler model that is also trained. In the context of a multispectral image dataset, the first layer captures meaningful features and patterns, as depicted in Fig. \ref{fig_4}. Additionally, the stacked architecture acts as a regularizer, preventing overfitting by forcing the model to learn more generalized features. Each layer acts as a feature extractor, reducing the data's dimensionality and encouraging the model to learn more generalizable features. Stacked autoencoders can extract features from the input data for use in other machine learning models, thereby improving their performance \citep{zhou2019learning}.

Unlike linear dimensionality reduction methods such as PCA, autoencoders do not aim to preserve specific attributes such as distance or topology. In scenarios where the relationships between input features are deep and nonlinear \citep{zhang2018automated}, traditional dimensionality reduction methods often fail to yield satisfactory results \citep{zhong2021advances}. Recognizing this limitation, the stacked autoencoder was developed, as the canonical autoencoder alone may struggle to address the nonlinearity inherent in many applications \citep{li2021distributed}. The architecture of a stacked autoencoder is entirely learned from the data, ensuring that the model adapts to the data's characteristics without the need for manual selection of nonlinear functions. In many cases, determining whether the relationship between spectral bands in a remote sensing dataset and target characteristics is linear or nonlinear can be challenging. The stacked autoencoder addresses both global and local characteristics within the dataset while reducing its dimensionality \citep{zhang2018local}.

\subsection{Clustering}

Unsupervised machine learning identifies patterns and relationships within the data autonomously without the need for labelled data. This is particularly useful for tasks such as clustering, anomaly detection, and association \citep{yadav2019study,awange2020hybrid}. Clustering, a key technique in unsupervised learning, involves grouping a set of objects in such a way that objects in the same group (or cluster) are more similar to each other than to those in other groups \citep{jain1999data}. Popular clustering methods include hierarchical clustering, DBSCAN (Density-Based Spatial Clustering of Applications with Noise), and $k$-means clustering \citep{jain2010data}. $k$-means clustering is one of the most widely used due to its simplicity and efficiency. It partitions the data into $k$ clusters, where each data point belongs to the cluster with the nearest mean, serving as the cluster's centroid \citep{sakthivel2021conspectus}.

$k$-means clustering has numerous applications, particularly in image processing and geological mapping \citep{shirmard2022review}. $k$-means is often used for image processing (computer vision) tasks such as image segmentation, where the goal is to partition an image into meaningful regions for easier analysis \citep{sakthivel2021conspectus}. In the case of remote sensing data, the algorithm works by treating pixel values as data points and grouping similar pixels into clusters, thereby simplifying the image into distinct segments \citep{selim1984kmeans}. In geological mapping, $k$-means can be instrumental in categorizing different landforms or mineral compositions based on remote sensing data. Geologists can identify and map various geological features, such as rock types, by clustering spectral data from satellite images \citep{shirmard2022review}. This application not only enhances the accuracy of geological surveys but also aids in resource exploration and environmental monitoring. The versatility and effectiveness of $k$-means make it a valuable tool in both image processing and remote sensing-based geological mapping, enabling the extraction of meaningful patterns and insights from complex datasets.

\subsubsection{Elbow method}

The elbow method is a widely used heuristic for determining the optimal number of clusters in $k$-means clustering \citep{celebi2013comparative}. This method involves running $k$-means clustering for a range of $k$ values and plotting the resulting sum of squared errors (SSE), also known as the within-cluster sum of squares (WCSS). The SSE measures the compactness of the clusters, with lower values indicating more tightly packed clusters \citep{patel2022approaches}. As $k$ increases, SSE naturally decreases because clusters have fewer data points, making them more compact. The key idea of the elbow method is to identify the point where SSE improvement dramatically slows down, forming an elbow in the plot \citep{onumanyi2022autoelbow}. This point suggests a suitable trade-off between the number of clusters and the compactness of the clusters, representing the optimal $k$. The elbow method is popular due to its simplicity and intuitive graphical representation, making it accessible even to those without advanced statistical knowledge \citep{onumanyi2022autoelbow}.

Despite its advantages, the elbow method has several limitations. One major drawback is its reliance on visual interpretation, which can be subjective and may lead to different conclusions depending on the interpreter's judgement \citep{shi2021quantitative}. In some cases, the elbow in the SSE plot might not be clearly defined, making it difficult to pinpoint the optimal $k$. Additionally, the elbow method assumes that the best clustering solution is where the rate of SSE improvement slows down, but this might not always align with the actual structure of the data \citep{shi2021quantitative}. In the case of datasets with complex or overlapping cluster structures, the elbow method may not provide a clear or accurate determination of $k$. Moreover, the method does not account for the possibility of multiple valid clustering solutions, each potentially useful for different applications \citep{shi2021quantitative}. These disadvantages suggest that while the elbow method is a valuable tool, it is often best used in conjunction with other techniques to ensure a more robust determination of the optimal number of clusters. However, given the high number of pixels in the satellite images used in this study and the time-consuming process of calculating other statistics like the silhouette score \citep{rousseeuw1987silhouettes}, we rely on the elbow method. We use the \textit{KElbowVisualizer} from the \textit{yellowbrick} Python library \citep{bengfort2018yellowbrick} to determine the optimal number of clusters. This module employs the Kneedle algorithm \citep{satopaa2011finding}, which detects the elbow point by identifying the maximum curvature in the plot of WCSS. The algorithm normalizes the WCSS values, calculates the difference between these values and a linear approximation, and identifies the point with the maximum difference as the elbow. This approach provides a reliable and automated method for determining the optimal $k$.

\subsection{Framework}

\begin{figure*}
    \begin{center}
    \includegraphics[width=\textwidth]{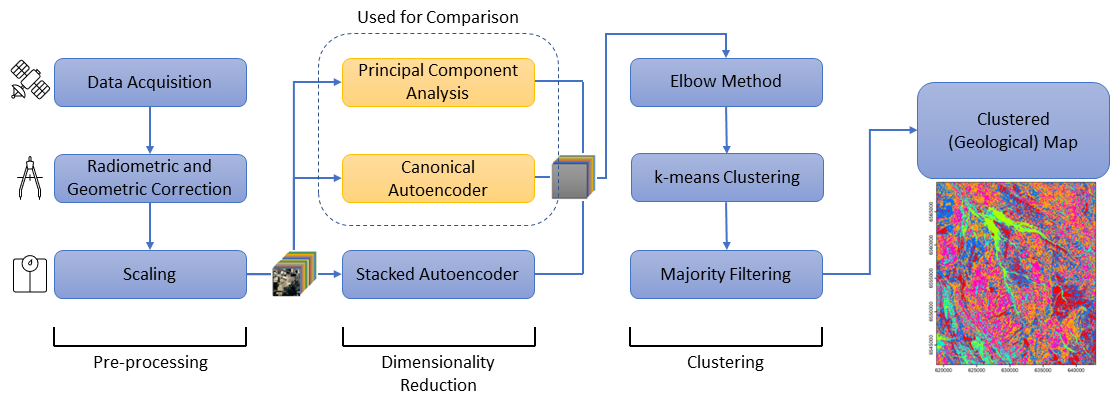}
    \caption{Machine learning framework for creating geological maps using the integration of the dimensionality reduction (PCA, canonical autoencoder, and stacked autoencoder) methods and the $k$-means clustering.}
    \end{center}
    \label{fig_5}
\end{figure*}

We need to adopt a multidisciplinary approach that integrates specialized domain-specific knowledge to implement deep learning models for geological mapping. In this study, we leverage stacked autoencoders and fine-tune their weights by training the model on various datasets to enhance its effectiveness in accurately identifying geological features. We highlighted earlier that autoencoders can be more robust to outliers and noisy data than PCA, as they can learn to ignore or suppress noisy features during training \citep{vincent2008extracting}. Fig. \ref{fig_5} illustrates the first step: acquiring multispectral data from Landsat 8, ASTER, and Sentinel-2, followed by necessary radiometric and geometric corrections and data scaling, which constitute the pre-processing stage. Next, we implement various dimensionality reduction methods, including PCA, canonical autoencoders, and stacked autoencoders, to create a compressed dataset (latent features) with reduced dimensions. Each pixel of this compressed data is considered a collection of non-spatial spectral observations by spectral classifiers. Subsequently, we apply clustering to the compressed data from these dimensionality reduction methods. We determine the optimal number of clusters using the elbow method to implement $k$-means and generate clustered maps representing geological features. Finally, we generate the clustered maps and interpret the results from a geological perspective.

We compare different dimensionality reduction methods and data types after implementing the clustering phase. We utilize metrics that include the Davies-Bouldin index and the variance ratio criterion (Calinski-Harabasz index), which can be calculated without labelled data \citep{patel2022approaches}. The Calinski-Harabasz index measures the ratio of between-cluster variation to within-cluster variance, with higher values indicating better-defined clusters. Therefore, the ideal number of clusters corresponds to maps with a high Calinski-Harabasz index, whereas lower values of the Davies-Bouldin index indicate better model performance.

\citet{mittal2022comprehensive} reported that neighbouring pixels in images are likely to belong to the same cluster due to spatial correlation, which aids in the clustering process. Geological units typically have regional distributions in space \citep{suchet2003worldwide}, meaning that adjacent pixels with similar properties are likely to belong to the same geological unit. However, each group of pixels may be close to a neighbouring cluster, which can create confusion. Applying a majority filter to the clustered map can improve its accuracy and consistency by removing outliers and inconsistencies, as well as the noise associated with the data. Since autoencoders map the data distribution to a normal distribution for input images without filtering (including noise), mapping without filtering will not be optimal and may result in blurred maps with fewer sharp boundaries. The majority filter can enhance the accuracy and reliability of geological maps, leading to a better understanding of an area's geology. In this study, we apply a majority filter with a kernel size of $7 \times 7$ to the clustered maps, assigning spurious pixels within a large single class to that class. The centre pixel in the kernel is replaced with the class value that the majority of the pixels in the kernel possess \citep{kantakumar2015multitemporal}.

If we remove majority filtering from the framework, anomalous pixels may indicate either remote sensing errors or specific non-homogeneities in geological structures. Anomalous pixels resulting from remote sensing errors can be identified through cross-validation with ground truth data and other remote sensing datasets. We can analyze these anomalies without filtering to improve the accuracy and reliability of the remote sensing data. Methods such as statistical outlier detection or comparison with high-resolution imagery can differentiate between true geological anomalies and errors. The anomalous pixels may also indicate genuine non-homogeneities within geological structures, which can provide significant geological insights. In this case, detailed field studies and additional sampling might be necessary to understand the underlying causes of these non-homogeneities.

\subsection{Implementation}

We implement the framework shown in Fig. \ref{fig_5} using the Python programming language and various machine learning packages, including Keras\footnote{Keras: \url{https://keras.io/api/}}, which facilitate the efficient implementation and execution of deep learning models and streamline the experimentation process. Implementing deep learning models typically requires setting several hyperparameters, and we experiment with different values to achieve the most accurate results.

In the case of PCA, we select the principal components that preserve $90\%$ of the total variance of the input spectral bands, resulting in a different number of components for each data type. The architecture of the canonical autoencoders includes one hidden layer and ten iterations (epochs). We use the rectified linear unit (ReLU) \citep{nair2010rectified} as the activation function for the hidden layers to account for non-linearity and the sigmoid activation function for the output layer. We use the Adam optimizer \citep{kingma2014adam} and mean squared error (MSE) loss function for model training.

The architecture of the stacked autoencoders used in this study comprises two autoencoders with one hidden layer for each. The optimizer, loss, and activation functions for the hidden and output layers are the same as those used for the canonical autoencoders. We provide the code implementation in a GitHub repository\footnote{\url{https://github.com/sydney-machine-learning/autoencoders_remotesensing}} and execute the experiments using an 11th Gen Intel(R) Core(TM) i7-11700 @ 2.50GHz CPU.

\section{Results}

\begin{figure*}
    \centering
    \includegraphics[width=\textwidth]{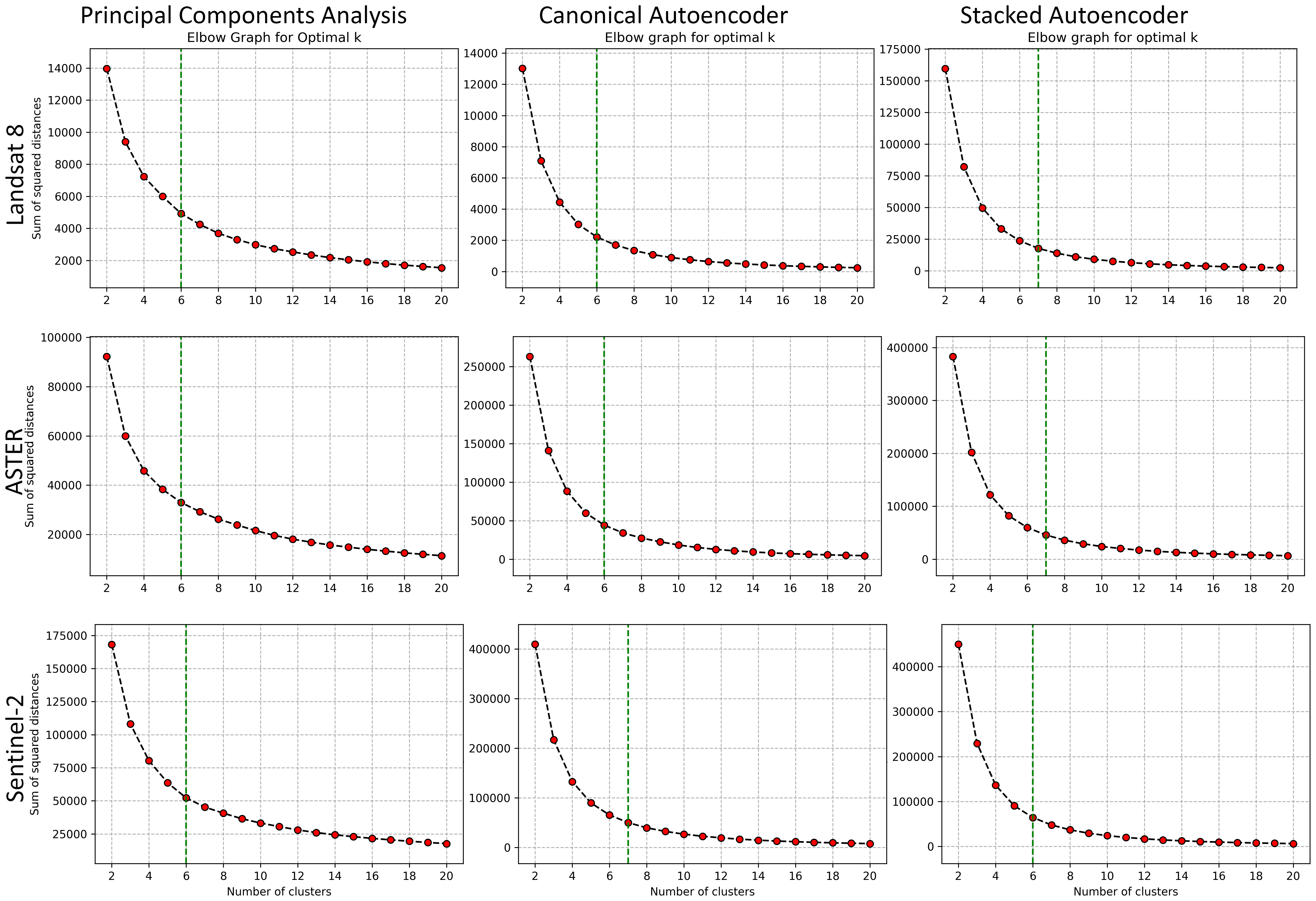}
    \caption{Elbow plots obtained for each pair of the data types and dimensionality reduction methods. The green dashed lines represent the optimal number of clusters.}
    \label{fig_6}
\end{figure*}

We apply the proposed framework to Landsat 8, ASTER, and Sentinel-2 multispectral data from the Mutawintji region in western New South Wales, Australia. Fig. \ref{fig_6} displays the elbow graphs used to determine the optimal number of clusters for different data types and dimensionality reduction method pairs. In these plots, the optimal $k$ for $k$-means clustering is indicated by a green dashed line, and the black line represents the sum of squared distances to the cluster centres. The point of maximum curvature in the elbow plots marks the optimal $k$ \citep{yuan2019research}. Table \ref{table_1} summarizes the optimal numbers of clusters, suggesting that six or seven major geological units in the study area exhibit specific spectral characteristics depending on the data type and dimensionality reduction method.

In addition to PCA, we train canonical and stacked autoencoders and calculate the reconstruction loss, demonstrating that the loss of information/features after dimensionality reduction is minimal. The features learned from the canonical and stacked autoencoders are then used to cluster the remote sensing data. We observe that the reconstruction loss stabilizes after a few epochs. Several metrics are available to evaluate the efficiency of clustering without labelled data, including the Silhouette coefficient, Calinski-Harabasz index, and Davies-Bouldin score \citep{patel2022approaches}. Due to the large number of pixels, calculating the Silhouette coefficient is time-consuming on a standard computer and is impractical. Therefore, we evaluate model performance using the Calinski-Harabasz and Davies-Bouldin scores \citep{renjith2020pragmatic,gao2021generalized}, as shown in Table \ref{table_2}. A higher Calinski-Harabasz score (first column for each method in Table \ref{table_2}) indicates better-defined clusters, while a lower Davies-Bouldin score suggests more efficient clustering results. It is noteworthy that we conducted additional analyses to quantify the impact of varying numbers of clusters ($k$) on our results. Specifically, we calculated the Calinski-Harabasz and Davies-Bouldin scores for $k$ values of 5, 6, 7, and 8 across all pairs of remote sensing data and dimensionality reduction methods. We found that the $k$ values determined by the Kneedle algorithm yielded the best results according to both the Calinski-Harabasz and Davies-Bouldin scores.

\begin{figure*}
\centering
\includegraphics[width=\textwidth]{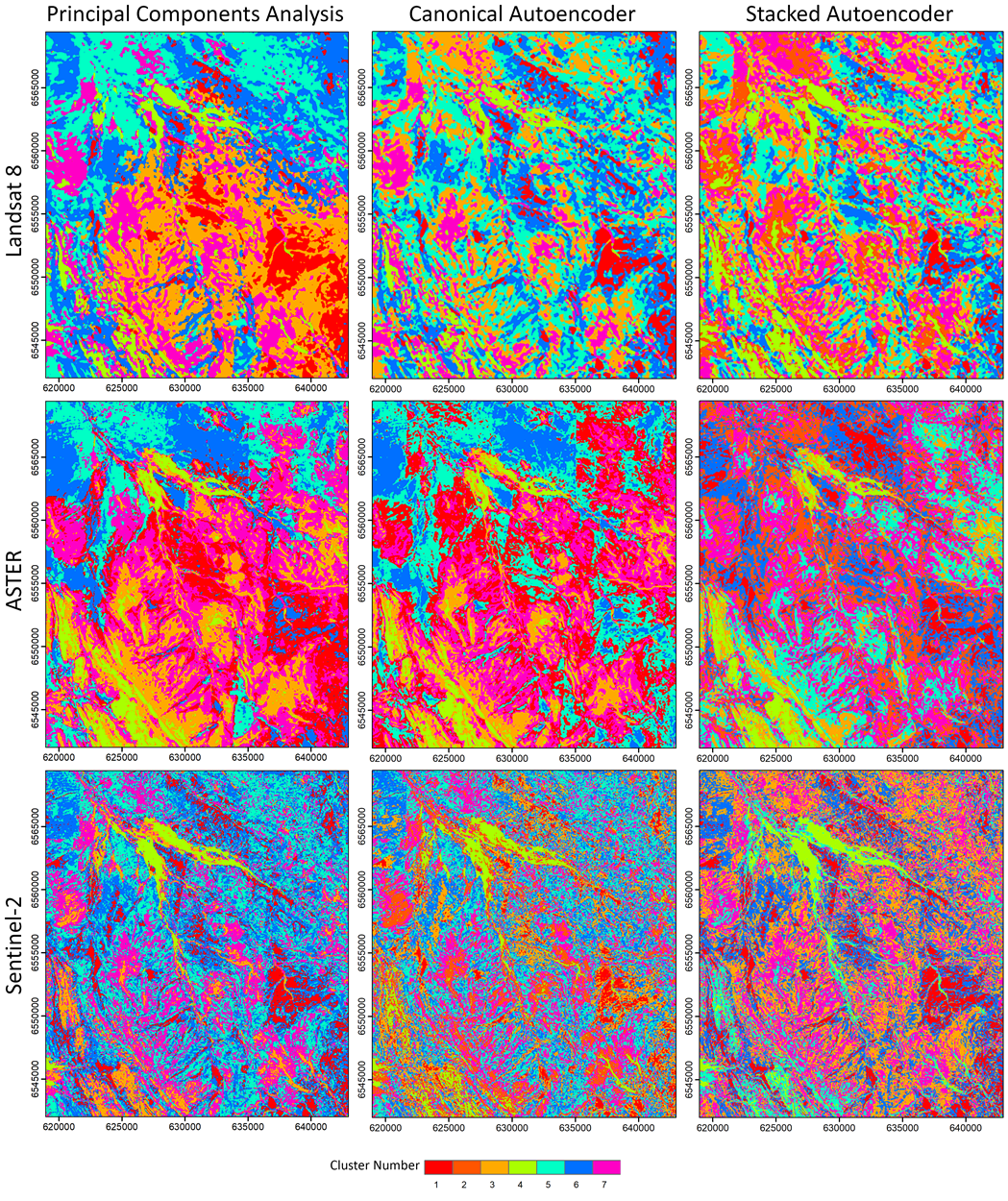}
\caption{Clustered maps of the study area obtained by different pairs of data types and dimensionality reduction methods. Each distinct coloured region in the maps represents a unique geological unit on the ground. We find that PCA (first column) generally results in less detailed geological unit differentiation than autoencoder-based methods, reflecting its limitations in capturing complex non-linear relationships in the data. Sentinel-2 data (last row) shows the best cluster compactness and separation performance, indicating a more distinct geological unit classification.}
\label{fig_7}
\end{figure*}

\begin{figure*}
\centering
\includegraphics[width=\textwidth]{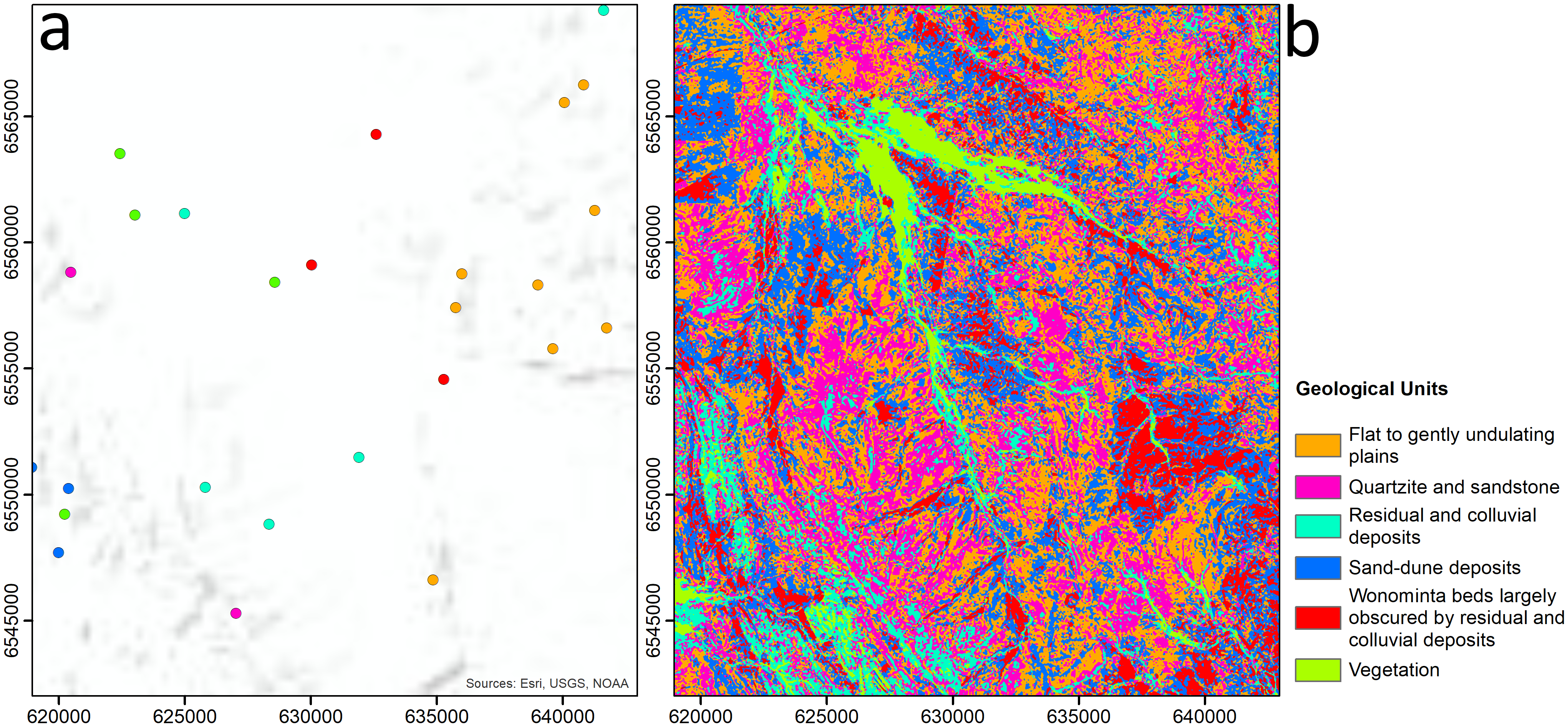}
\caption{a) Rock samples available from across the study area through MinView; b) The clustered map interpreted by implementing the proposed framework and applying stacked autoencoders to the Sentinel-2 dataset.}
\label{fig_8}
\end{figure*}

Fig. \ref{fig_7} presents the clustered maps obtained by applying the framework to various pairs of remote sensing data and dimensionality reduction methods. We observe that geological maps generated using PCA followed by $k$-means clustering differ from those produced by canonical and stacked autoencoders. The maps generated using stacked autoencoders on Landsat 8 and ASTER, as well as canonical autoencoders on Sentinel-2, consist of seven clusters, while the others have six clusters. In addition to the previously mentioned metrics, we use 30 rock samples and associated information provided by the Geological Survey of NSW\footnote{\url{https://minview.geoscience.nsw.gov.au}} as ground truth data. We calculate the overall accuracy for each pair of dimensionality reduction methods and remote sensing data by dividing the correctly clustered samples by the total number of samples (Table \ref{table_3}). Given that the optimal number of clusters varies among data types and dimensionality reduction methods, we simplify the rock types and categorize them into six (Fig. \ref{fig_8}(a)) or seven classes according to Table \ref{table_1}.

\begin{table*}[htbp]
\centering
\caption{Optimal number of clusters determined by the elbow graphs for different pairs of data types and dimensionality reduction methods.}
\label{table_1}
\begin{tabular}{|l|c|c|c|}
\hline
\multicolumn{1}{|c|}{Data Type/Method} & PCA & Canonical Autoencoder & Stacked Autoencoder \\ \hline
Landsat 8                              & 6   & 6                     & 7                   \\ \hline
ASTER                                  & 6   & 6                     & 7                   \\ \hline
Sentinel 2                             & 6   & 7                     & 6                   \\ \hline
\end{tabular}
\end{table*}

\begin{table*}[htbp]
\centering
\caption{Calinski-Harabasz and Davies-Bouldin scores calculated for different pairs of data types and dimensionality reduction methods.}
\label{table_2}
\resizebox{\textwidth}{!}{%
\begin{tabular}{|l|cc|cc|cc|}
\hline
\multicolumn{1}{|c|}{\multirow{2}{*}{Data Type/Method}} & \multicolumn{2}{c|}{PCA}                                & \multicolumn{2}{c|}{Canonical Autoencoder}              & \multicolumn{2}{c|}{Stacked Autoencoder}                \\ \cline{2-7} 
\multicolumn{1}{|c|}{}                                  & \multicolumn{1}{c|}{Calinski-Harabasz} & Davies-Bouldin & \multicolumn{1}{c|}{Calinski-Harabasz} & Davies-Bouldin & \multicolumn{1}{c|}{Calinski-Harabasz} & Davies-Bouldin \\ \hline
Landsat 8                                               & \multicolumn{1}{c|}{725156}            & 0.848          & \multicolumn{1}{c|}{2049661}           & 0.538          & \multicolumn{1}{c|}{2928417}           & 0.520          \\ \hline
ASTER                                                   & \multicolumn{1}{c|}{3203478}           & 0.882          & \multicolumn{1}{c|}{8940866}           & 0.534          & \multicolumn{1}{c|}{10204767}          & 0.533          \\ \hline
Sentinel 2                                              & \multicolumn{1}{c|}{7975444}           & 0.799          & \multicolumn{1}{c|}{20868268}          & 0.530          & \multicolumn{1}{c|}{22433325}          & 0.525          \\ \hline
\end{tabular}%
}
\end{table*}

\begin{table*}[htbp]
\centering
\caption{Overall accuracy of different pairs of data types and dimensionality reduction methods based on ground truth data (rock samples).}
\label{table_3}
\begin{tabular}{|l|c|c|c|}
\hline
\multicolumn{1}{|c|}{Data Type/Method} & PCA   & Canonical Autoencoder & Stacked Autoencoder \\ \hline
Landsat 8                              & 0.767 & 0.800                 & 0.866               \\ \hline
ASTER                                  & 0.833 & 0.833                 & 0.900               \\ \hline
Sentinel 2                             & 0.800 & 0.833                 & 0.900               \\ \hline
\end{tabular}
\end{table*}

We compare the clustered maps obtained from our framework (Fig. \ref{fig_7}) to evaluate the efficiency of different dimensionality reduction methods in terms of noise removal, data compression, and geological unit discrimination. We use three metrics to compare these methods. According to Table \ref{table_2}, stacked autoencoders achieve larger Calinski-Harabasz scores and lower Davies-Bouldin scores for each data type, indicating that the clusters generated using the latent vectors of stacked autoencoders are more separable compared to principal components and the latent vectors obtained by canonical autoencoders. We also observe that as spatial resolution increases, both scores improve, with Sentinel-2 yielding the highest scores due to its superior spatial resolution. Among all data types and dimensionality reduction methods, stacked autoencoders applied to Sentinel-2 data provide the best results.

Each uniquely coloured region in the maps shown in Fig. \ref{fig_7} represents a distinct geological unit clustered using a pair of dimensionality reduction methods and data types. By comparing these clustered maps with the geological map shown in Fig. \ref{fig_1}(b), it is evident that Sentinel-2 data produce a more detailed clustered map than other data types and conventional approaches for mapping geological units. Sentinel-2 data uniquely identifies vegetation and assigns a separate cluster to relevant pixels. Additionally, the comparison of dimensionality reduction methods reveals that only stacked autoencoders accurately differentiate between various sedimentary units in the south and southeast of the study area. Utilizing high-resolution remote sensing data such as Sentinel-2, combined with non-linear dimensionality reduction methods like stacked autoencoders, enhances the signal-to-noise ratio of input features to clustering algorithms, resulting in more detailed and accurate geological maps than conventional approaches like field surveys.

The metrics reported in Table \ref{table_2} do not consider ground truth or labelled data, relying solely on input features and assigned labels for each observation or pixel. The overall accuracy presented in Table \ref{table_3} incorporates the adaptation of collected rock samples from the study area shown in Fig. \ref{fig_8}(a) with the clustered maps, providing a more reliable metric for determining the best approach for generating a clustered or geological map. According to Table \ref{table_3}, employing stacked autoencoders on Sentinel-2 data yields the highest accuracy, indicating that the majority of rock samples have been assigned to the correct cluster. Consequently, the map generated using Sentinel-2 and stacked autoencoders is interpreted as the geological map shown in Fig. \ref{fig_7}(b). This map identifies five different geological units plus vegetation, effectively distinguishing between different sedimentary units in the study area, such as sandstones, residual and colluvial deposits, and undulating plains.

\section{Discussion}

The results presented in this study provide insights into the efficacy of different dimensionality reduction methods combined with various remote sensing data types for geological mapping. The evaluation metrics offer quantitative validation of the proposed machine learning framework for geological mapping via dimensionality reduction and clustering. According to Table \ref{table_2}, the Calinski-Harabasz score indicates the separability of clusters generated by PCA and autoencoders, with stacked autoencoders consistently yielding higher scores across all data types compared to PCA. This suggests that the non-linear nature of autoencoders facilitates better discrimination between geological units, particularly with Sentinel-2 data, which benefits from higher spatial resolution. The Davies-Bouldin score, which assesses cluster compactness and separation, reveals similar findings, with the stacked autoencoder on Sentinel-2 data showing the best performance. This highlights the importance of considering spectral characteristics and data suitability for specific models when evaluating clustering performance in remote sensing applications.

The visual examination of clustered maps (Fig. \ref{fig_7}) further demonstrates the advantages of using Sentinel-2 data with stacked autoencoders. The resultant map provides detailed and accurate delineation for geological mapping. Moreover, stacked autoencoders exhibit superior discrimination capabilities, especially for different sedimentary units in specific regions of the study area. Including ground truth data enhances the assessment, with overall accuracy serving as a robust metric. The results underscore the effectiveness of stacked autoencoders on Sentinel-2 data, yielding the highest accuracy and ensuring the proper assignment of rock samples to clusters (Table \ref{table_3}). Consequently, leveraging Sentinel-2 data and stacked autoencoders produces a detailed geological map of the study area, successfully discriminating between various geological units.

Although autoencoders are more computationally intensive to train and may require extensive hyperparameter tuning, the findings demonstrate that the proposed framework yields compelling results for geological mapping applications using multispectral data without labelled data. However, the effectiveness of the proposed approach should be validated across diverse geological terrains to assess its broader applicability. Additionally, preprocessing steps applied to remote sensing data, such as atmospheric correction, radiometric calibration, or geometric registration, can significantly impact the quality and accuracy of the derived geological maps and should be carefully considered.

The choice of the number of clusters ($k$) introduces uncertainties in the creation of clustered maps, which are interpreted as geological maps. An incorrect $k$ value can lead to either over-segmentation or under-segmentation of geological units, affecting the accuracy and reliability of the map. Over-segmentation might result in an excessive number of clusters, misrepresenting homogenous geological units as multiple distinct entities. Conversely, under-segmentation may group distinct geological units together, masking important geological variations.

Additionally, the low number of rock samples can further introduce uncertainties. Limited samples may not adequately capture the variability of geological units, leading to less accurate cluster definitions. This insufficiency can result in clustered maps that do not accurately reflect the true geological diversity of the area. These uncertainties have significant implications for geological map interpretations. Variations in $k$ values and sample sizes can lead to different geological unit definitions, affecting exploration and decision-making processes. For example, misidentified units might lead to incorrect assumptions about mineral resources, structural stability, or environmental conditions. Therefore, careful consideration of $k$ values and efforts to obtain a sufficient number of representative rock samples are crucial. Employing robust validation techniques and cross-referencing with existing geological data can help mitigate these uncertainties, leading to more accurate and reliable geological maps.

In our study, the stationarity assumptions overlook the dynamic nature of geological features (e.g., due to landslides, earthquakes, seasonal erosion, and human activities), necessitating methods to incorporate temporal dynamics for more accurate mapping. Reliance on subjective ground truth data introduces uncertainty, which can be mitigated by integrating multiple sources and using consensus-based approaches. Although advanced techniques improve discrimination, efforts are needed to enhance the interpretability of the resulting maps. Other dimensionality reduction methods, such as manifold learning \citep{izenman2012introduction,pless2009survey}, can be used to extend our framework. This study illustrated the potential of unsupervised feature learning methods in feature extraction, motivating large-scale applications using hyperspectral datasets. Moreover, incorporating ancillary data sources, such as geological maps, digital elevation models, and hydrological data, could enrich the analysis and improve the discrimination of geological units. The framework can be extended with novel clustering methods, such as spectral and hierarchical clustering \citep{saxena2017review} and Gaussian mixture models, which have shown promising results in remote sensing and image-based data processing \citep{deo2024reefcoreseg,barve2023reef}.

\section{Conclusions}

We presented a framework that combined stacked autoencoders with $k$-means clustering to generate geological maps. By applying various pairs of remote sensing data types and dimensionality reduction methods, we created input features for the $k$-means algorithm, resulting in automated geological maps for the Mutawintji region in NSW, Australia. Our investigation revealed that the combination integration of stacked autoencoders with Sentinel-2 data yields the
highest spatial resolution and accuracy when compared to other combinations. The stacked autoencoders demonstrated
  to be highly effective for dimensionality reduction and feature learning, enabling better the extraction of  complex and hierarchical representations of the input data when compared to canonical autoencoders and PCA.
Our findings reveal that the integration of stacked autoencoders with Sentinel-2 data yields the highest spatial resolution and accuracy. The stacked autoencoders proved to be highly effective for dimensionality reduction and feature learning, enabling the extraction of more complex and hierarchical representations of the input data compared to canonical autoencoders and PCA.

The flexibility of our framework allows for further enhancements with the incorporation of novel dimensionality reduction and clustering methods. This adaptability ensures that our approach can evolve alongside advancements in remote sensing and machine learning techniques, making it a robust tool for geological mapping in various regions and contexts.

\section*{Code and Data Availability}
The Python code and datasets used to implement the framework are available at \url{https://github.com/sydney-machine-learning/autoencoders_remotesensing}.

\bibliographystyle{jasr-model5-names}
\biboptions{authoryear}
\bibliography{References}

\begin{thebibliography}{108}
\expandafter\ifx\csname natexlab\endcsname\relax\def\natexlab#1{#1}\fi
\ifx\xfnm\relax \def\xfnm[#1]{\unskip,\space#1}\fi
%Type = Article

\bibitem[{Abdi \& Williams(2010)}]{abdi2010principal}
\bibinfo{author}{Abdi, H.},  \& \bibinfo{author}{Williams, L.~J.}
  (\bibinfo{year}{2010}).
\newblock \bibinfo{title}{Principal component analysis}.
\newblock {\it \bibinfo{journal}{Wiley Interdisciplinary Reviews: Computational
  Statistics}\/},  {\it \bibinfo{volume}{2}\/}\bibinfo{issue}{(4)},
  \bibinfo{pages}{433--459}.
%Type = Article

\bibitem[{Akanbi et~al.(2024)Akanbi, Bhuvanagiri, Barcelos, Nihar,
  Gonzalez~Hernandez, Yarus \& French}]{akanbi2024integrating}
\bibinfo{author}{Akanbi, O.~D.}, \bibinfo{author}{Bhuvanagiri, D.~C.},
  \bibinfo{author}{Barcelos, E.~I.} et~al. (\bibinfo{year}{2024}).
\newblock \bibinfo{title}{Integrating multiscale geospatial analysis for
  monitoring crop growth, nutrient distribution, and hydrological dynamics in
  large-scale agricultural systems}.
\newblock {\it \bibinfo{journal}{Journal of Geovisualization and Spatial
  Analysis}\/},  {\it \bibinfo{volume}{8}\/}\bibinfo{issue}{(1)},
  \bibinfo{pages}{9}. \DOIprefix\doi{10.1007/s41651-023-00164-y}.
%Type = Article

\bibitem[{Amare et~al.(2024)Amare, Demissie, Beza \& Erena}]{amare2024impacts}
\bibinfo{author}{Amare, M.~T.}, \bibinfo{author}{Demissie, S.~T.},
  \bibinfo{author}{Beza, S.~A.} et~al. (\bibinfo{year}{2024}).
\newblock \bibinfo{title}{Impacts of land use/land cover changes on the
  hydrology of the fafan catchment ethiopia}.
\newblock {\it \bibinfo{journal}{Journal of Geovisualization and Spatial
  Analysis}\/},  {\it \bibinfo{volume}{8}\/}\bibinfo{issue}{(1)},
  \bibinfo{pages}{10}. \DOIprefix\doi{10.1007/s41651-024-00172-6}.
%Type = Article

\bibitem[{Amiotte~Suchet et~al.(2003)Amiotte~Suchet, Probst \&
  Ludwig}]{suchet2003worldwide}
\bibinfo{author}{Amiotte~Suchet, P.}, \bibinfo{author}{Probst, J.-L.},  \&
  \bibinfo{author}{Ludwig, W.} (\bibinfo{year}{2003}).
\newblock \bibinfo{title}{Worldwide distribution of continental rock lithology:
  Implications for the atmospheric/soil co2 uptake by continental weathering
  and alkalinity river transport to the oceans}.
\newblock {\it \bibinfo{journal}{Global Biogeochemical Cycles}\/},  {\it
  \bibinfo{volume}{17}\/}\bibinfo{issue}{(2)}.
%Type = Article

\bibitem[{dos Anjos et~al.(2021)dos Anjos, Avila, Vasconcelos, Pereira~Neta,
  Medeiros, Evsukoff, Surmas \& Landau}]{dos2021deep}
\bibinfo{author}{dos Anjos, C.~E.}, \bibinfo{author}{Avila, M.~R.},
  \bibinfo{author}{Vasconcelos, A.~G.} et~al. (\bibinfo{year}{2021}).
\newblock \bibinfo{title}{Deep learning for lithological classification of
  carbonate rock micro-ct images}.
\newblock {\it \bibinfo{journal}{Computational Geosciences}\/},  {\it
  \bibinfo{volume}{25}\/}\bibinfo{issue}{(3)}, \bibinfo{pages}{971--983}.
%Type = Article

\bibitem[{Asadzadeh \& de~Souza~Filho(2016)}]{asadzadeh2016review}
\bibinfo{author}{Asadzadeh, S.},  \& \bibinfo{author}{de~Souza~Filho, C.~R.}
  (\bibinfo{year}{2016}).
\newblock \bibinfo{title}{A review on spectral processing methods for
  geological remote sensing}.
\newblock {\it \bibinfo{journal}{International Journal of Applied Earth
  Observation and Geoinformation}\/},  {\it \bibinfo{volume}{47}\/},
  \bibinfo{pages}{69--90}.
%Type = Book

\bibitem[{Awange et~al.(2020)Awange, Pal{\'a}ncz \&
  V{\"o}lgyesi}]{awange2020hybrid}
\bibinfo{author}{Awange, J.}, \bibinfo{author}{Pal{\'a}ncz, B.},  \&
  \bibinfo{author}{V{\"o}lgyesi, L.} (\bibinfo{year}{2020}).
\newblock {\it \bibinfo{title}{Hybrid Imaging and Visualization. Employing
  Machine Learning with Mathematica-Python}\/}.
\newblock \bibinfo{publisher}{Springer}.
%Type = Article

\bibitem[{Bachri et~al.(2019)Bachri, Hakdaoui, Raji, Teodoro \&
  Benbouziane}]{bachri2019machine}
\bibinfo{author}{Bachri, I.}, \bibinfo{author}{Hakdaoui, M.},
  \bibinfo{author}{Raji, M.} et~al. (\bibinfo{year}{2019}).
\newblock \bibinfo{title}{Machine learning algorithms for automatic
  lithological mapping using remote sensing data: A case study from souk arbaa
  sahel, sidi ifni inlier, western anti-atlas, morocco}.
\newblock {\it \bibinfo{journal}{ISPRS International Journal of
  Geo-Information}\/},  {\it \bibinfo{volume}{8}\/}\bibinfo{issue}{(6)},
  \bibinfo{pages}{248}.
%Type = Article

\bibitem[{Bandyopadhyay et~al.(2007)Bandyopadhyay, Maulik \&
  Mukhopadhyay}]{bandyopadhyay2007multiobjective}
\bibinfo{author}{Bandyopadhyay, S.}, \bibinfo{author}{Maulik, U.},  \&
  \bibinfo{author}{Mukhopadhyay, A.} (\bibinfo{year}{2007}).
\newblock \bibinfo{title}{Multiobjective genetic clustering for pixel
  classification in remote sensing imagery}.
\newblock {\it \bibinfo{journal}{IEEE transactions on Geoscience and Remote
  Sensing}\/},  {\it \bibinfo{volume}{45}\/}\bibinfo{issue}{(5)},
  \bibinfo{pages}{1506--1511}.
%Type = Article

\bibitem[{Barovich \& Hand(2008)}]{barovich2008tectonic}
\bibinfo{author}{Barovich, K.},  \& \bibinfo{author}{Hand, M.}
  (\bibinfo{year}{2008}).
\newblock \bibinfo{title}{Tectonic setting and provenance of the
  paleoproterozoic willyama supergroup, curnamona province, australia:
  Geochemical and nd isotopic constraints on contrasting source terrain
  components}.
\newblock {\it \bibinfo{journal}{Precambrian Research}\/},  {\it
  \bibinfo{volume}{166}\/}\bibinfo{issue}{(1)}, \bibinfo{pages}{318--337}.
%Type = Article

\bibitem[{Barve et~al.(2023)Barve, Webster \& Chandra}]{barve2023reef}
\bibinfo{author}{Barve, S.}, \bibinfo{author}{Webster, J.~M.},  \&
  \bibinfo{author}{Chandra, R.} (\bibinfo{year}{2023}).
\newblock \bibinfo{title}{Reef-insight: A framework for reef habitat mapping
  with clustering methods using remote sensing}.
\newblock {\it \bibinfo{journal}{Information}\/},  {\it
  \bibinfo{volume}{14}\/}\bibinfo{issue}{(7)}, \bibinfo{pages}{373}.
%Type = Article

\bibitem[{Bedini(2009)}]{bedini2009mapping}
\bibinfo{author}{Bedini, E.} (\bibinfo{year}{2009}).
\newblock \bibinfo{title}{Mapping lithology of the sarfartoq carbonatite
  complex, southern west greenland, using hymap imaging spectrometer data}.
\newblock {\it \bibinfo{journal}{Remote Sensing of Environment}\/},  {\it
  \bibinfo{volume}{113}\/}\bibinfo{issue}{(6)}, \bibinfo{pages}{1208--1219}.
%Type = Article

\bibitem[{Behnia et~al.(2012)Behnia, Harris, Rainbird, Williamson \&
  Sheshpari}]{behnia2012remote}
\bibinfo{author}{Behnia, P.}, \bibinfo{author}{Harris, J.},
  \bibinfo{author}{Rainbird, R.} et~al. (\bibinfo{year}{2012}).
\newblock \bibinfo{title}{Remote predictive mapping of bedrock geology using
  image classification of landsat and spot data, western minto inlier, victoria
  island, northwest territories, canada}.
\newblock {\it \bibinfo{journal}{International journal of remote sensing}\/},
  {\it \bibinfo{volume}{33}\/}\bibinfo{issue}{(21)},
  \bibinfo{pages}{6876--6903}.
%Type = Misc

\bibitem[{Bengfort et~al.(2018)Bengfort, Bilbro, Danielsen, Gray, {McIntyre},
  Roman, Poh et~al.}]{bengfort2018yellowbrick}
\bibinfo{author}{Bengfort, B.}, \bibinfo{author}{Bilbro, R.},
  \bibinfo{author}{Danielsen, N.} et~al. (\bibinfo{year}{2018}).
\newblock \bibinfo{title}{Yellowbrick}.
\newblock \URLprefix \url{http://www.scikit-yb.org/en/latest/}.
  \DOIprefix\doi{10.5281/zenodo.1206264}.
%Type = Article

\bibitem[{Bentahar et~al.(2020)Bentahar, Raji \&
  Si~Mhamdi}]{bentahar2020fracture}
\bibinfo{author}{Bentahar, I.}, \bibinfo{author}{Raji, M.},  \&
  \bibinfo{author}{Si~Mhamdi, H.} (\bibinfo{year}{2020}).
\newblock \bibinfo{title}{Fracture network mapping using landsat-8 oli,
  sentinel-2a, aster, and aster-gdem data, in the rich area (central high
  atlas, morocco)}.
\newblock {\it \bibinfo{journal}{Arabian Journal of Geosciences}\/},  {\it
  \bibinfo{volume}{13}\/}\bibinfo{issue}{(16)}, \bibinfo{pages}{768}.
%Type = Article

\bibitem[{Bo et~al.(2009)Bo, Ding, Li, Di \& Zhu}]{bo2009mean}
\bibinfo{author}{Bo, S.}, \bibinfo{author}{Ding, L.}, \bibinfo{author}{Li, H.}
  et~al. (\bibinfo{year}{2009}).
\newblock \bibinfo{title}{Mean shift-based clustering analysis of multispectral
  remote sensing imagery}.
\newblock {\it \bibinfo{journal}{International Journal of Remote Sensing}\/},
  {\it \bibinfo{volume}{30}\/}\bibinfo{issue}{(4)}, \bibinfo{pages}{817--827}.
%Type = Article

\bibitem[{Bouslihim et~al.(2022)Bouslihim, Kharrou, Miftah, Attou, Bouchaou \&
  Chehbouni}]{bouslihim2022comparing}
\bibinfo{author}{Bouslihim, Y.}, \bibinfo{author}{Kharrou, M.~H.},
  \bibinfo{author}{Miftah, A.} et~al. (\bibinfo{year}{2022}).
\newblock \bibinfo{title}{Comparing pan-sharpened landsat-9 and sentinel-2 for
  land-use classification using machine learning classifiers}.
\newblock {\it \bibinfo{journal}{Journal of Geovisualization and Spatial
  Analysis}\/},  {\it \bibinfo{volume}{6}\/}\bibinfo{issue}{(2)},
  \bibinfo{pages}{35}. \DOIprefix\doi{10.1007/s41651-022-00130-0}.
%Type = Inbook

\bibitem[{Bruzzone \& Demir(2014)}]{bruzzone2014review}
\bibinfo{author}{Bruzzone, L.},  \& \bibinfo{author}{Demir, B.}
  (\bibinfo{year}{2014}).
\newblock \bibinfo{title}{A review of modern approaches to classification of
  remote sensing data}.
\newblock In {\it \bibinfo{booktitle}{Land Use and Land Cover Mapping in
  Europe: Practices {\&} Trends}\/} (pp. \bibinfo{pages}{127--143}).
\newblock \bibinfo{publisher}{Springer Netherlands}.
%Type = Inproceedings

\bibitem[{Calvin(2018)}]{calvin2018band}
\bibinfo{author}{Calvin, W.~M.} (\bibinfo{year}{2018}).
\newblock \bibinfo{title}{Band parameterization for imaging spectrometer
  systems: Lessons learned from crism at mars}.
\newblock In {\it \bibinfo{booktitle}{IEEE International Geoscience and Remote
  Sensing Symposium}\/} (pp. \bibinfo{pages}{8356--8358}).
%Type = Article

\bibitem[{Carneiro et~al.(2012)Carneiro, Fraser, Cr{\'o}sta, Silva \&
  Barros}]{carneiro2012semiautomated}
\bibinfo{author}{Carneiro, C. d.~C.}, \bibinfo{author}{Fraser, S.~J.},
  \bibinfo{author}{Cr{\'o}sta, A.~P.} et~al. (\bibinfo{year}{2012}).
\newblock \bibinfo{title}{Semiautomated geologic mapping using self-organizing
  maps and airborne geophysics in the brazilian amazon}.
\newblock {\it \bibinfo{journal}{Geophysics}\/},  {\it
  \bibinfo{volume}{77}\/}\bibinfo{issue}{(4)}, \bibinfo{pages}{K17--K24}.
%Type = Article

\bibitem[{Celebi et~al.(2013)Celebi, Kingravi \& Vela}]{celebi2013comparative}
\bibinfo{author}{Celebi, M.~E.}, \bibinfo{author}{Kingravi, H.~A.},  \&
  \bibinfo{author}{Vela, P.~A.} (\bibinfo{year}{2013}).
\newblock \bibinfo{title}{A comparative study of efficient initialization
  methods for the k-means clustering algorithm}.
\newblock {\it \bibinfo{journal}{Expert Systems with Applications}\/},  {\it
  \bibinfo{volume}{40}\/}\bibinfo{issue}{(1)}, \bibinfo{pages}{200--210}.
  \DOIprefix\doi{10.1016/j.eswa.2012.07.021}.
%Type = Article

\bibitem[{Chen et~al.(2010)Chen, Warner \& Campagna}]{chen2010integrating}
\bibinfo{author}{Chen, X.}, \bibinfo{author}{Warner, T.~A.},  \&
  \bibinfo{author}{Campagna, D.~J.} (\bibinfo{year}{2010}).
\newblock \bibinfo{title}{Integrating visible, near-infrared and short-wave
  infrared hyperspectral and multispectral thermal imagery for geological
  mapping at cuprite, nevada: a rule-based system}.
\newblock {\it \bibinfo{journal}{International Journal of Remote Sensing}\/},
  {\it \bibinfo{volume}{31}\/}\bibinfo{issue}{(7)},
  \bibinfo{pages}{1733--1752}.
%Type = Article

\bibitem[{Clark et~al.(2003)Clark, Swayze, Livo, Kokaly, Sutley, Dalton,
  McDougal \& Gent}]{clark2003imaging}
\bibinfo{author}{Clark, R.~N.}, \bibinfo{author}{Swayze, G.~A.},
  \bibinfo{author}{Livo, K.~E.} et~al. (\bibinfo{year}{2003}).
\newblock \bibinfo{title}{Imaging spectroscopy: Earth and planetary remote
  sensing with the usgs tetracorder and expert systems}.
\newblock {\it \bibinfo{journal}{Journal of Geophysical Research: Planets}\/},
  {\it \bibinfo{volume}{108}\/}\bibinfo{issue}{(E12)}.
%Type = Article

\bibitem[{Comon(1994)}]{comon1994independent}
\bibinfo{author}{Comon, P.} (\bibinfo{year}{1994}).
\newblock \bibinfo{title}{Independent component analysis, a new concept?}
\newblock {\it \bibinfo{journal}{Signal Processing}\/},  {\it
  \bibinfo{volume}{36}\/}\bibinfo{issue}{(3)}, \bibinfo{pages}{287--314}.
%Type = Article

\bibitem[{Dai et~al.(2023)Dai, Cheng, Guo, Wang, Qu, Liu, Li, Lu, Wang, Zeng
  et~al.}]{dai2023optimization}
\bibinfo{author}{Dai, X.}, \bibinfo{author}{Cheng, J.}, \bibinfo{author}{Guo,
  S.} et~al. (\bibinfo{year}{2023}).
\newblock \bibinfo{title}{Optimization strategy of a stacked autoencoder and
  deep belief network in a hyperspectral remote-sensing image classification
  model}.
\newblock {\it \bibinfo{journal}{Discrete Dynamics in Nature and Society}\/},
  {\it \bibinfo{volume}{2023}\/}.
%Type = Article

\bibitem[{Davies \& Bouldin(1979)}]{davies1979cluster}
\bibinfo{author}{Davies, D.~L.},  \& \bibinfo{author}{Bouldin, D.~W.}
  (\bibinfo{year}{1979}).
\newblock \bibinfo{title}{A cluster separation measure}.
\newblock {\it \bibinfo{journal}{IEEE Transactions on Pattern Analysis and
  Machine Intelligence}\/},  {\it \bibinfo{volume}{1}\/}\bibinfo{issue}{(2)},
  \bibinfo{pages}{224--227}.
%Type = Article

\bibitem[{Deo et~al.(2024)Deo, Webster, Salles \& Chandra}]{deo2024reefcoreseg}
\bibinfo{author}{Deo, R.}, \bibinfo{author}{Webster, J.~M.},
  \bibinfo{author}{Salles, T.} et~al. (\bibinfo{year}{2024}).
\newblock \bibinfo{title}{Reefcoreseg: A clustering-based framework for
  multi-source data fusion for segmentation of reef drill cores}.
\newblock {\it \bibinfo{journal}{IEEE Access}\/},  {\it
  \bibinfo{volume}{12}\/}, \bibinfo{pages}{12164--12180}.
%Type = Article

\bibitem[{Dou et~al.(2024{\natexlab{a}})Dou, Huang, Han, Hou \&
  Zhang}]{dou2024time}
\bibinfo{author}{Dou, P.}, \bibinfo{author}{Huang, C.}, \bibinfo{author}{Han,
  W.} et~al. (\bibinfo{year}{2024}{\natexlab{a}}).
\newblock \bibinfo{title}{Time series remote sensing image classification using
  feature relationship learning}.
\newblock {\it \bibinfo{journal}{IEEE Transactions on Geoscience and Remote
  Sensing}\/},  {\it \bibinfo{volume}{62}\/}, \bibinfo{pages}{1--13}.
  \DOIprefix\doi{10.1109/TGRS.2024.3386171}.
%Type = Article

\bibitem[{Dou et~al.(2024{\natexlab{b}})Dou, Huang, Han, Hou, Zhang \&
  Gu}]{dou2024remote}
\bibinfo{author}{Dou, P.}, \bibinfo{author}{Huang, C.}, \bibinfo{author}{Han,
  W.} et~al. (\bibinfo{year}{2024}{\natexlab{b}}).
\newblock \bibinfo{title}{Remote sensing image classification using an ensemble
  framework without multiple classifiers}.
\newblock {\it \bibinfo{journal}{ISPRS Journal of Photogrammetry and Remote
  Sensing}\/},  {\it \bibinfo{volume}{208}\/}, \bibinfo{pages}{190--209}.
  \DOIprefix\doi{10.1016/j.isprsjprs.2023.12.012}.
%Type = Article

\bibitem[{Dou et~al.(2024{\natexlab{c}})Dou, Shen, Huang, Li, Mao \&
  Li}]{dou2024large}
\bibinfo{author}{Dou, P.}, \bibinfo{author}{Shen, H.}, \bibinfo{author}{Huang,
  C.} et~al. (\bibinfo{year}{2024}{\natexlab{c}}).
\newblock \bibinfo{title}{Large-scale land use/land cover extraction from
  landsat imagery using feature relationships matrix based deep-shallow
  learning}.
\newblock {\it \bibinfo{journal}{International Journal of Applied Earth
  Observation and Geoinformation}\/},  {\it \bibinfo{volume}{129}\/},
  \bibinfo{pages}{103866}. \DOIprefix\doi{10.1016/j.jag.2024.103866}.
%Type = Article

\bibitem[{Drusch et~al.(2012)Drusch, {Del Bello}, Carlier, Colin, Fernandez,
  Gascon, Hoersch, Isola, Laberinti, Martimort, Meygret, Spoto, Sy, Marchese \&
  Bargellini}]{drusch2012sentinel2}
\bibinfo{author}{Drusch, M.}, \bibinfo{author}{{Del Bello}, U.},
  \bibinfo{author}{Carlier, S.} et~al. (\bibinfo{year}{2012}).
\newblock \bibinfo{title}{Sentinel-2: Esa's optical high-resolution mission for
  gmes operational services}.
\newblock {\it \bibinfo{journal}{Remote Sensing of Environment}\/},  {\it
  \bibinfo{volume}{120}\/}, \bibinfo{pages}{25--36}.
%Type = Article

\bibitem[{Fan et~al.(2009)Fan, Han \& Wang}]{fan2009single}
\bibinfo{author}{Fan, J.}, \bibinfo{author}{Han, M.},  \&
  \bibinfo{author}{Wang, J.} (\bibinfo{year}{2009}).
\newblock \bibinfo{title}{Single point iterative weighted fuzzy c-means
  clustering algorithm for remote sensing image segmentation}.
\newblock {\it \bibinfo{journal}{Pattern Recognition}\/},  {\it
  \bibinfo{volume}{42}\/}\bibinfo{issue}{(11)}, \bibinfo{pages}{2527--2540}.
%Type = Article

\bibitem[{Forootan et~al.(2012)Forootan, Awange, Kusche, Heck \&
  Eicker}]{forootan2012independent}
\bibinfo{author}{Forootan, E.}, \bibinfo{author}{Awange, J.~L.},
  \bibinfo{author}{Kusche, J.} et~al. (\bibinfo{year}{2012}).
\newblock \bibinfo{title}{Independent patterns of water mass anomalies over
  australia from satellite data and models}.
\newblock {\it \bibinfo{journal}{Remote Sensing of Environment}\/},  {\it
  \bibinfo{volume}{124}\/}, \bibinfo{pages}{427--443}.
%Type = Article

\bibitem[{Galdames et~al.(2019)Galdames, Perez, Estévez \&
  Adams}]{galdames2019rock}
\bibinfo{author}{Galdames, F.~J.}, \bibinfo{author}{Perez, C.~A.},
  \bibinfo{author}{Estévez, P.~A.} et~al. (\bibinfo{year}{2019}).
\newblock \bibinfo{title}{Rock lithological classification by hyperspectral,
  range 3d and color images}.
\newblock {\it \bibinfo{journal}{Chemometrics and Intelligent Laboratory
  Systems}\/},  {\it \bibinfo{volume}{189}\/}, \bibinfo{pages}{138--148}.
  \DOIprefix\doi{10.1016/j.chemolab.2019.04.006}.
%Type = Inproceedings

\bibitem[{Gao et~al.(2021)Gao, Rasmussen, Kulits, Scheller, Greenberger \&
  Ehlmann}]{gao2021generalized}
\bibinfo{author}{Gao, A.~F.}, \bibinfo{author}{Rasmussen, B.},
  \bibinfo{author}{Kulits, P.} et~al. (\bibinfo{year}{2021}).
\newblock \bibinfo{title}{Generalized unsupervised clustering of hyperspectral
  images of geological targets in the near infrared}.
\newblock In {\it \bibinfo{booktitle}{Proceedings of the IEEE/CVF Conference on
  Computer Vision and Pattern Recognition}\/} (pp.
  \bibinfo{pages}{4294--4303}).
%Type = Article

\bibitem[{Gao et~al.(2017)Gao, Zhao, Jia, Liao \& Zhang}]{gao2017optimized}
\bibinfo{author}{Gao, L.}, \bibinfo{author}{Zhao, B.}, \bibinfo{author}{Jia,
  X.} et~al. (\bibinfo{year}{2017}).
\newblock \bibinfo{title}{Optimized kernel minimum noise fraction
  transformation for hyperspectral image classification}.
\newblock {\it \bibinfo{journal}{Remote sensing}\/},  {\it
  \bibinfo{volume}{9}\/}\bibinfo{issue}{(6)}, \bibinfo{pages}{548}.
%Type = Article

\bibitem[{Gewali et~al.(2018)Gewali, Monteiro \& Saber}]{gewali2018machine}
\bibinfo{author}{Gewali, U.~B.}, \bibinfo{author}{Monteiro, S.~T.},  \&
  \bibinfo{author}{Saber, E.} (\bibinfo{year}{2018}).
\newblock \bibinfo{title}{Machine learning based hyperspectral image analysis:
  A survey}.
\newblock {\it \bibinfo{journal}{arXiv}\/},  {\it
  \bibinfo{volume}{1802.08701}\/}.
%Type = Article

\bibitem[{Ghosh et~al.(2011)Ghosh, Mishra \& Ghosh}]{ghosh2011fuzzy}
\bibinfo{author}{Ghosh, A.}, \bibinfo{author}{Mishra, N.~S.},  \&
  \bibinfo{author}{Ghosh, S.} (\bibinfo{year}{2011}).
\newblock \bibinfo{title}{Fuzzy clustering algorithms for unsupervised change
  detection in remote sensing images}.
\newblock {\it \bibinfo{journal}{Information Sciences}\/},  {\it
  \bibinfo{volume}{181}\/}\bibinfo{issue}{(4)}, \bibinfo{pages}{699--715}.
%Type = Inproceedings

\bibitem[{Guo et~al.(2023)Guo, Zhang, Fu \& Yang}]{guo2023gis}
\bibinfo{author}{Guo, Y.}, \bibinfo{author}{Zhang, S.}, \bibinfo{author}{Fu,
  C.} et~al. (\bibinfo{year}{2023}).
\newblock \bibinfo{title}{Gis-based mineral prospectivity mapping: A systematic
  study on machine learning at hezuo-meiwu district, gansu province}.
\newblock In \bibinfo{editor}{Y.~Wang} (Ed.), {\it
  \bibinfo{booktitle}{International Conference on Geographic Information and
  Remote Sensing Technology (GIRST 2022)}\/} (p. \bibinfo{pages}{125523A}).
\newblock volume \bibinfo{volume}{12552}.
\newblock \DOIprefix\doi{10.1117/12.2667272}.
%Type = Article

\bibitem[{Hajaj et~al.(2024)Hajaj, {El Harti}, Pour, Jellouli, Adiri \&
  Hashim}]{hajaj2024review}
\bibinfo{author}{Hajaj, S.}, \bibinfo{author}{{El Harti}, A.},
  \bibinfo{author}{Pour, A.~B.} et~al. (\bibinfo{year}{2024}).
\newblock \bibinfo{title}{A review on hyperspectral imagery application for
  lithological mapping and mineral prospecting: Machine learning techniques and
  future prospects}.
\newblock {\it \bibinfo{journal}{Remote Sensing Applications: Society and
  Environment}\/},  {\it \bibinfo{volume}{35}\/}, \bibinfo{pages}{101218}.
  \DOIprefix\doi{10.1016/j.rsase.2024.101218}.
%Type = Article

\bibitem[{Hashim et~al.(2013)Hashim, Ahmad, Johari \&
  Pour}]{hashim2013automatic}
\bibinfo{author}{Hashim, M.}, \bibinfo{author}{Ahmad, S.},
  \bibinfo{author}{Johari, M. A.~M.} et~al. (\bibinfo{year}{2013}).
\newblock \bibinfo{title}{Automatic lineament extraction in a heavily vegetated
  region using landsat enhanced thematic mapper (etm+) imagery}.
\newblock {\it \bibinfo{journal}{Advances in Space Research}\/},  {\it
  \bibinfo{volume}{51}\/}\bibinfo{issue}{(5)}, \bibinfo{pages}{874--890}.
  \DOIprefix\doi{10.1016/j.asr.2012.10.004}.
%Type = Article

\bibitem[{Hewson et~al.(2005)Hewson, Cudahy, Mizuhiko, Ueda \&
  Mauger}]{hewson2005seamless}
\bibinfo{author}{Hewson, R.}, \bibinfo{author}{Cudahy, T.},
  \bibinfo{author}{Mizuhiko, S.} et~al. (\bibinfo{year}{2005}).
\newblock \bibinfo{title}{Seamless geological map generation using aster in the
  broken hill-curnamona province of australia}.
\newblock {\it \bibinfo{journal}{Remote Sensing of Environment}\/},  {\it
  \bibinfo{volume}{99}\/}\bibinfo{issue}{(1)}, \bibinfo{pages}{159--172}.
%Type = Article

\bibitem[{Izenman(2012)}]{izenman2012introduction}
\bibinfo{author}{Izenman, A.~J.} (\bibinfo{year}{2012}).
\newblock \bibinfo{title}{Introduction to manifold learning}.
\newblock {\it \bibinfo{journal}{Wiley Interdisciplinary Reviews: Computational
  Statistics}\/},  {\it \bibinfo{volume}{4}\/}\bibinfo{issue}{(5)},
  \bibinfo{pages}{439--446}.
%Type = Article

\bibitem[{Jain(2010)}]{jain2010data}
\bibinfo{author}{Jain, A.~K.} (\bibinfo{year}{2010}).
\newblock \bibinfo{title}{Data clustering: 50 years beyond k-means}.
\newblock {\it \bibinfo{journal}{Pattern Recognition Letters}\/},  {\it
  \bibinfo{volume}{31}\/}\bibinfo{issue}{(8)}, \bibinfo{pages}{651--666}.
  \DOIprefix\doi{10.1016/j.patrec.2009.09.011}.
%Type = Article

\bibitem[{Jain et~al.(1999)Jain, Murty \& Flynn}]{jain1999data}
\bibinfo{author}{Jain, A.~K.}, \bibinfo{author}{Murty, M.~N.},  \&
  \bibinfo{author}{Flynn, P.~J.} (\bibinfo{year}{1999}).
\newblock \bibinfo{title}{Data clustering: A review}.
\newblock {\it \bibinfo{journal}{ACM Computing Surveys}\/},  {\it
  \bibinfo{volume}{31}\/}\bibinfo{issue}{(3)}, \bibinfo{pages}{264--–323}.
  \DOIprefix\doi{10.1145/331499.331504}.
%Type = Article

\bibitem[{Kantakumar \& Neelamsetti(2015)}]{kantakumar2015multitemporal}
\bibinfo{author}{Kantakumar, L.~N.},  \& \bibinfo{author}{Neelamsetti, P.}
  (\bibinfo{year}{2015}).
\newblock \bibinfo{title}{Multi-temporal land use classification using hybrid
  approach}.
\newblock {\it \bibinfo{journal}{The Egyptian Journal of Remote Sensing and
  Space Science}\/},  {\it \bibinfo{volume}{18}\/}\bibinfo{issue}{(2)},
  \bibinfo{pages}{289--295}.
%Type = Article

\bibitem[{Kingma \& Ba(2014)}]{kingma2014adam}
\bibinfo{author}{Kingma, D.~P.},  \& \bibinfo{author}{Ba, J.}
  (\bibinfo{year}{2014}).
\newblock \bibinfo{title}{Adam: A method for stochastic optimization}.
\newblock {\it \bibinfo{journal}{arXiv}\/},  {\it
  \bibinfo{volume}{arXiv:1412.6980}\/}.
%Type = Article

\bibitem[{Kingma \& Welling(2019)}]{kingma2019introduction}
\bibinfo{author}{Kingma, D.~P.},  \& \bibinfo{author}{Welling, M.}
  (\bibinfo{year}{2019}).
\newblock \bibinfo{title}{An introduction to variational autoencoders}.
\newblock {\it \bibinfo{journal}{arXiv}\/},  {\it
  \bibinfo{volume}{arXiv:1906.02691}\/}.
%Type = Article

\bibitem[{Kramer(1992)}]{kramer1992autoassociative}
\bibinfo{author}{Kramer, M.~A.} (\bibinfo{year}{1992}).
\newblock \bibinfo{title}{Autoassociative neural networks}.
\newblock {\it \bibinfo{journal}{Computers \& chemical engineering}\/},  {\it
  \bibinfo{volume}{16}\/}\bibinfo{issue}{(4)}, \bibinfo{pages}{313--328}.
%Type = Article

\bibitem[{Li et~al.(2023)Li, Pei \& Li}]{li2023comprehensive}
\bibinfo{author}{Li, P.}, \bibinfo{author}{Pei, Y.},  \& \bibinfo{author}{Li,
  J.} (\bibinfo{year}{2023}).
\newblock \bibinfo{title}{A comprehensive survey on design and application of
  autoencoder in deep learning}.
\newblock {\it \bibinfo{journal}{Applied Soft Computing}\/},  {\it
  \bibinfo{volume}{138}\/}, \bibinfo{pages}{110176}.
%Type = Article

\bibitem[{Li et~al.(2021)Li, Tian, Jiang \& Yan}]{li2021distributed}
\bibinfo{author}{Li, Z.}, \bibinfo{author}{Tian, L.}, \bibinfo{author}{Jiang,
  Q.} et~al. (\bibinfo{year}{2021}).
\newblock \bibinfo{title}{Distributed-ensemble stacked autoencoder model for
  non-linear process monitoring}.
\newblock {\it \bibinfo{journal}{Information Sciences}\/},  {\it
  \bibinfo{volume}{542}\/}, \bibinfo{pages}{302--316}.
%Type = Inproceedings

\bibitem[{L{\'o}pez-Fandi{\~n}o et~al.(2018)L{\'o}pez-Fandi{\~n}o, Garea, Heras
  \& Arg{\"u}ello}]{lopez2018stacked}
\bibinfo{author}{L{\'o}pez-Fandi{\~n}o, J.}, \bibinfo{author}{Garea, A.~S.},
  \bibinfo{author}{Heras, D.~B.} et~al. (\bibinfo{year}{2018}).
\newblock \bibinfo{title}{Stacked autoencoders for multiclass change detection
  in hyperspectral images}.
\newblock In {\it \bibinfo{booktitle}{IEEE International Geoscience and Remote
  Sensing Symposium}\/} (pp. \bibinfo{pages}{1906--1909}).
%Type = Article

\bibitem[{Lu et~al.(2021)Lu, Yang \& Meng}]{lu2021lithology}
\bibinfo{author}{Lu, Y.}, \bibinfo{author}{Yang, C.},  \&
  \bibinfo{author}{Meng, Z.} (\bibinfo{year}{2021}).
\newblock \bibinfo{title}{Lithology discrimination using sentinel-1 dual-pol
  data and srtm data}.
\newblock {\it \bibinfo{journal}{Remote Sensing}\/},  {\it
  \bibinfo{volume}{13}\/}\bibinfo{issue}{(7)}, \bibinfo{pages}{1280}.
%Type = Article

\bibitem[{Lv et~al.(2017)Lv, Han \& Qiu}]{lv2017remote}
\bibinfo{author}{Lv, F.}, \bibinfo{author}{Han, M.},  \& \bibinfo{author}{Qiu,
  T.} (\bibinfo{year}{2017}).
\newblock \bibinfo{title}{Remote sensing image classification based on ensemble
  extreme learning machine with stacked autoencoder}.
\newblock {\it \bibinfo{journal}{IEEE Access}\/},  {\it \bibinfo{volume}{5}\/},
  \bibinfo{pages}{9021--9031}.
%Type = Article

\bibitem[{Ma{\'c}kiewicz \& Ratajczak(1993)}]{mackiewicz1993principal}
\bibinfo{author}{Ma{\'c}kiewicz, A.},  \& \bibinfo{author}{Ratajczak, W.}
  (\bibinfo{year}{1993}).
\newblock \bibinfo{title}{Principal components analysis (pca)}.
\newblock {\it \bibinfo{journal}{Computers \& Geosciences}\/},  {\it
  \bibinfo{volume}{19}\/}\bibinfo{issue}{(3)}, \bibinfo{pages}{303--342}.
%Type = Article

\bibitem[{Masoudi et~al.(2024)Masoudi, Richards \& Tan}]{masoudi2024assessment}
\bibinfo{author}{Masoudi, M.}, \bibinfo{author}{Richards, D.~R.},  \&
  \bibinfo{author}{Tan, P.~Y.} (\bibinfo{year}{2024}).
\newblock \bibinfo{title}{Assessment of the influence of spatial scale and type
  of land cover on urban landscape pattern analysis using landscape metrics}.
\newblock {\it \bibinfo{journal}{Journal of Geovisualization and Spatial
  Analysis}\/},  {\it \bibinfo{volume}{8}\/}\bibinfo{issue}{(1)},
  \bibinfo{pages}{8}. \DOIprefix\doi{10.1007/s41651-024-00170-8}.
%Type = Article

\bibitem[{Mittal et~al.(2022)Mittal, Pandey, Saraswat, Kumar, Pal \&
  Modwel}]{mittal2022comprehensive}
\bibinfo{author}{Mittal, H.}, \bibinfo{author}{Pandey, A.~C.},
  \bibinfo{author}{Saraswat, M.} et~al. (\bibinfo{year}{2022}).
\newblock \bibinfo{title}{A comprehensive survey of image segmentation:
  clustering methods, performance parameters, and benchmark datasets}.
\newblock {\it \bibinfo{journal}{Multimedia Tools and Applications}\/},  (pp.
  \bibinfo{pages}{1--26}).
%Type = Inproceedings

\bibitem[{Nair \& Hinton(2010)}]{nair2010rectified}
\bibinfo{author}{Nair, V.},  \& \bibinfo{author}{Hinton, G.~E.}
  (\bibinfo{year}{2010}).
\newblock \bibinfo{title}{Rectified linear units improve restricted boltzmann
  machines}.
\newblock In {\it \bibinfo{booktitle}{27th International Conference on Machine
  Learning}\/} (pp. \bibinfo{pages}{807--814}).
%Type = Article

\bibitem[{Nalepa et~al.(2020)Nalepa, Myller, Imai, Honda, Takeda \&
  Antoniak}]{nalepa2020unsupervised}
\bibinfo{author}{Nalepa, J.}, \bibinfo{author}{Myller, M.},
  \bibinfo{author}{Imai, Y.} et~al. (\bibinfo{year}{2020}).
\newblock \bibinfo{title}{Unsupervised segmentation of hyperspectral images
  using 3-d convolutional autoencoders}.
\newblock {\it \bibinfo{journal}{IEEE Geoscience and Remote Sensing
  Letters}\/},  {\it \bibinfo{volume}{17}\/}\bibinfo{issue}{(11)},
  \bibinfo{pages}{1948--1952}.
%Type = Article

\bibitem[{Nielsen(2010)}]{nielsen2010kernel}
\bibinfo{author}{Nielsen, A.~A.} (\bibinfo{year}{2010}).
\newblock \bibinfo{title}{Kernel maximum autocorrelation factor and minimum
  noise fraction transformations}.
\newblock {\it \bibinfo{journal}{IEEE Transactions on Image Processing}\/},
  {\it \bibinfo{volume}{20}\/}\bibinfo{issue}{(3)}, \bibinfo{pages}{612--624}.
%Type = Article

\bibitem[{Omran et~al.(2007)Omran, Engelbrecht \& Salman}]{omran2007overview}
\bibinfo{author}{Omran, M.~G.}, \bibinfo{author}{Engelbrecht, A.~P.},  \&
  \bibinfo{author}{Salman, A.} (\bibinfo{year}{2007}).
\newblock \bibinfo{title}{An overview of clustering methods}.
\newblock {\it \bibinfo{journal}{Intelligent Data Analysis}\/},  {\it
  \bibinfo{volume}{11}\/}\bibinfo{issue}{(6)}, \bibinfo{pages}{583--605}.
%Type = Article

\bibitem[{Onumanyi et~al.(2022)Onumanyi, Molokomme, Isaac \&
  Abu-Mahfouz}]{onumanyi2022autoelbow}
\bibinfo{author}{Onumanyi, A.~J.}, \bibinfo{author}{Molokomme, D.~N.},
  \bibinfo{author}{Isaac, S.~J.} et~al. (\bibinfo{year}{2022}).
\newblock \bibinfo{title}{Autoelbow: An automatic elbow detection method for
  estimating the number of clusters in a dataset}.
\newblock {\it \bibinfo{journal}{Applied Sciences}\/},  {\it
  \bibinfo{volume}{12}\/}\bibinfo{issue}{(15)}.
  \DOIprefix\doi{10.3390/app12157515}.
%Type = Inproceedings

\bibitem[{{\"O}zdemir et~al.(2014){\"O}zdemir, Gedik \&
  {\c{C}}etin}]{ozdemir2014hyperspectral}
\bibinfo{author}{{\"O}zdemir, A. O.~B.}, \bibinfo{author}{Gedik, B.~E.},  \&
  \bibinfo{author}{{\c{C}}etin, C. Y.~Y.} (\bibinfo{year}{2014}).
\newblock \bibinfo{title}{Hyperspectral classification using stacked
  autoencoders with deep learning}.
\newblock In {\it \bibinfo{booktitle}{6th Workshop on Hyperspectral Image and
  Signal Processing: Evolution in Remote Sensing (WHISPERS)}\/} (pp.
  \bibinfo{pages}{1--4}).
%Type = Article

\bibitem[{Pal et~al.(2020)Pal, Rasmussen \& Porwal}]{pal2020optimized}
\bibinfo{author}{Pal, M.}, \bibinfo{author}{Rasmussen, T.},  \&
  \bibinfo{author}{Porwal, A.} (\bibinfo{year}{2020}).
\newblock \bibinfo{title}{Optimized lithological mapping from multispectral and
  hyperspectral remote sensing images using fused multi-classifiers}.
\newblock {\it \bibinfo{journal}{Remote Sensing}\/},  {\it
  \bibinfo{volume}{12}\/}\bibinfo{issue}{(1)}, \bibinfo{pages}{177}.
%Type = Inproceedings

\bibitem[{Patel et~al.(2022)Patel, Sivaiah \& Patel}]{patel2022approaches}
\bibinfo{author}{Patel, P.}, \bibinfo{author}{Sivaiah, B.},  \&
  \bibinfo{author}{Patel, R.} (\bibinfo{year}{2022}).
\newblock \bibinfo{title}{Approaches for finding optimal number of clusters
  using k-means and agglomerative hierarchical clustering techniques}.
\newblock In {\it \bibinfo{booktitle}{International Conference on Intelligent
  Controller and Computing for Smart Power}\/} (pp. \bibinfo{pages}{1--6}).
%Type = Article

\bibitem[{Pless \& Souvenir(2009)}]{pless2009survey}
\bibinfo{author}{Pless, R.},  \& \bibinfo{author}{Souvenir, R.}
  (\bibinfo{year}{2009}).
\newblock \bibinfo{title}{A survey of manifold learning for images}.
\newblock {\it \bibinfo{journal}{IPSJ Transactions on Computer Vision and
  Applications}\/},  {\it \bibinfo{volume}{1}\/}, \bibinfo{pages}{83--94}.
%Type = Article

\bibitem[{Pour et~al.(2018)Pour, Hashim, Park \& Hong}]{pour2018mapping}
\bibinfo{author}{Pour, A.~B.}, \bibinfo{author}{Hashim, M.},
  \bibinfo{author}{Park, Y.} et~al. (\bibinfo{year}{2018}).
\newblock \bibinfo{title}{Mapping alteration mineral zones and lithological
  units in antarctic regions using spectral bands of aster remote sensing
  data}.
\newblock {\it \bibinfo{journal}{Geocarto International}\/},  {\it
  \bibinfo{volume}{33}\/}\bibinfo{issue}{(12)}, \bibinfo{pages}{1281--1306}.
%Type = Article

\bibitem[{Protopapadakis et~al.(2021)Protopapadakis, Doulamis, Doulamis \&
  Maltezos}]{protopapadakis2021stacked}
\bibinfo{author}{Protopapadakis, E.}, \bibinfo{author}{Doulamis, A.},
  \bibinfo{author}{Doulamis, N.} et~al. (\bibinfo{year}{2021}).
\newblock \bibinfo{title}{Stacked autoencoders driven by semi-supervised
  learning for building extraction from near infrared remote sensing imagery}.
\newblock {\it \bibinfo{journal}{Remote Sensing}\/},  {\it
  \bibinfo{volume}{13}\/}\bibinfo{issue}{(3)}.
  \DOIprefix\doi{10.3390/rs13030371}.
%Type = Article

\bibitem[{Ran et~al.(2019)Ran, Xue, Zhang, Liu, Sang \& He}]{ran2019rock}
\bibinfo{author}{Ran, X.}, \bibinfo{author}{Xue, L.}, \bibinfo{author}{Zhang,
  Y.} et~al. (\bibinfo{year}{2019}).
\newblock \bibinfo{title}{Rock classification from field image patches analyzed
  using a deep convolutional neural network}.
\newblock {\it \bibinfo{journal}{Mathematics}\/},  {\it
  \bibinfo{volume}{7}\/}\bibinfo{issue}{(8)}, \bibinfo{pages}{755}.
%Type = Incollection

\bibitem[{Renjith et~al.(2020)Renjith, Sreekumar \&
  Jathavedan}]{renjith2020pragmatic}
\bibinfo{author}{Renjith, S.}, \bibinfo{author}{Sreekumar, A.},  \&
  \bibinfo{author}{Jathavedan, M.} (\bibinfo{year}{2020}).
\newblock \bibinfo{title}{Pragmatic evaluation of the impact of dimensionality
  reduction in the performance of clustering algorithms}.
\newblock In {\it \bibinfo{booktitle}{Advances in Electrical and Computer
  Technologies}\/} (pp. \bibinfo{pages}{499--512}).
\newblock \bibinfo{publisher}{Springer}.
%Type = Book

\bibitem[{Richards \& Richards(1999)}]{richards1999remote}
\bibinfo{author}{Richards, J.~A.},  \& \bibinfo{author}{Richards, J.}
  (\bibinfo{year}{1999}).
\newblock {\it \bibinfo{title}{Remote sensing digital image analysis}\/}
  volume~\bibinfo{volume}{3}.
\newblock \bibinfo{publisher}{Springer}.
%Type = Article

\bibitem[{Rodarmel \& Shan(2002)}]{rodarmel2002principal}
\bibinfo{author}{Rodarmel, C.},  \& \bibinfo{author}{Shan, J.}
  (\bibinfo{year}{2002}).
\newblock \bibinfo{title}{Principal component analysis for hyperspectral image
  classification}.
\newblock {\it \bibinfo{journal}{Surveying and Land Information Science}\/},
  {\it \bibinfo{volume}{62}\/}\bibinfo{issue}{(2)}, \bibinfo{pages}{115--122}.
%Type = Article

\bibitem[{Rousseeuw(1987)}]{rousseeuw1987silhouettes}
\bibinfo{author}{Rousseeuw, P.~J.} (\bibinfo{year}{1987}).
\newblock \bibinfo{title}{Silhouettes: A graphical aid to the interpretation
  and validation of cluster analysis}.
\newblock {\it \bibinfo{journal}{Journal of Computational and Applied
  Mathematics}\/},  {\it \bibinfo{volume}{20}\/}, \bibinfo{pages}{53--65}.
  \DOIprefix\doi{10.1016/0377-0427(87)90125-7}.
%Type = Article

\bibitem[{Rowan \& Mars(2003)}]{rowan2003lithologic}
\bibinfo{author}{Rowan, L.~C.},  \& \bibinfo{author}{Mars, J.~C.}
  (\bibinfo{year}{2003}).
\newblock \bibinfo{title}{Lithologic mapping in the mountain pass, california
  area using advanced spaceborne thermal emission and reflection radiometer
  (aster) data}.
\newblock {\it \bibinfo{journal}{Remote Sensing of Environment}\/},  {\it
  \bibinfo{volume}{84}\/}\bibinfo{issue}{(3)}, \bibinfo{pages}{350--366}.
%Type = Article

\bibitem[{Sagi \& Rokach(2018)}]{sagi2018ensemble}
\bibinfo{author}{Sagi, O.},  \& \bibinfo{author}{Rokach, L.}
  (\bibinfo{year}{2018}).
\newblock \bibinfo{title}{Ensemble learning: A survey}.
\newblock {\it \bibinfo{journal}{Wiley Interdisciplinary Reviews: Data Mining
  and Knowledge Discovery}\/},  {\it
  \bibinfo{volume}{8}\/}\bibinfo{issue}{(4)}, \bibinfo{pages}{e1249}.
%Type = Article

\bibitem[{Sahoo \& Jha(2017)}]{sahoo2017pattern}
\bibinfo{author}{Sahoo, S.},  \& \bibinfo{author}{Jha, M.~K.}
  (\bibinfo{year}{2017}).
\newblock \bibinfo{title}{Pattern recognition in lithology classification:
  modeling using neural networks, self-organizing maps and genetic algorithms}.
\newblock {\it \bibinfo{journal}{Hydrogeology journal}\/},  {\it
  \bibinfo{volume}{25}\/}\bibinfo{issue}{(2)}, \bibinfo{pages}{311--330}.
%Type = Article

\bibitem[{Sakthivel et~al.(2021)Sakthivel, Jyothi, Susila \&
  Sheela}]{sakthivel2021conspectus}
\bibinfo{author}{Sakthivel, U.}, \bibinfo{author}{Jyothi, A.},
  \bibinfo{author}{Susila, N.} et~al. (\bibinfo{year}{2021}).
\newblock \bibinfo{title}{Conspectus of k-means clustering algorithm}.
\newblock {\it \bibinfo{journal}{Applied Learning Algorithms for Intelligent
  IoT}\/},  (pp. \bibinfo{pages}{193--213}).
  \DOIprefix\doi{10.1201/9781003119838-9}.
%Type = Inproceedings

\bibitem[{Satopaa et~al.(2011)Satopaa, Albrecht, Irwin \&
  Raghavan}]{satopaa2011finding}
\bibinfo{author}{Satopaa, V.}, \bibinfo{author}{Albrecht, J.},
  \bibinfo{author}{Irwin, D.} et~al. (\bibinfo{year}{2011}).
\newblock \bibinfo{title}{Finding a ``kneedle'' in a haystack: Detecting knee
  points in system behavior}.
\newblock In {\it \bibinfo{booktitle}{31st International Conference on
  Distributed Computing Systems Workshops}\/} (pp. \bibinfo{pages}{166--171}).
\newblock \DOIprefix\doi{10.1109/ICDCSW.2011.20}.
%Type = Article

\bibitem[{Saxena et~al.(2017)Saxena, Prasad, Gupta, Bharill, Patel, Tiwari, Er,
  Ding \& Lin}]{saxena2017review}
\bibinfo{author}{Saxena, A.}, \bibinfo{author}{Prasad, M.},
  \bibinfo{author}{Gupta, A.} et~al. (\bibinfo{year}{2017}).
\newblock \bibinfo{title}{A review of clustering techniques and developments}.
\newblock {\it \bibinfo{journal}{Neurocomputing}\/},  {\it
  \bibinfo{volume}{267}\/}, \bibinfo{pages}{664--681}.
%Type = Article

\bibitem[{Selim \& Ismail(1984)}]{selim1984kmeans}
\bibinfo{author}{Selim, S.~Z.},  \& \bibinfo{author}{Ismail, M.~A.}
  (\bibinfo{year}{1984}).
\newblock \bibinfo{title}{K-means-type algorithms: A generalized convergence
  theorem and characterization of local optimality}.
\newblock {\it \bibinfo{journal}{IEEE Transactions on Pattern Analysis and
  Machine Intelligence}\/},  {\it
  \bibinfo{volume}{PAMI-6}\/}\bibinfo{issue}{(1)}, \bibinfo{pages}{81--87}.
  \DOIprefix\doi{10.1109/TPAMI.1984.4767478}.
%Type = Article

\bibitem[{Sgavetti et~al.(2006)Sgavetti, Pompilio \&
  Meli}]{sgavetti2006reflectance}
\bibinfo{author}{Sgavetti, M.}, \bibinfo{author}{Pompilio, L.},  \&
  \bibinfo{author}{Meli, S.} (\bibinfo{year}{2006}).
\newblock \bibinfo{title}{Reflectance spectroscopy (0.3–2.5 µm) at various
  scales for bulk-rock identification}.
\newblock {\it \bibinfo{journal}{Geosphere}\/},  {\it
  \bibinfo{volume}{2}\/}\bibinfo{issue}{(3)}, \bibinfo{pages}{142--160}.
%Type = Article

\bibitem[{Shebl \& Árpád Csámer(2021)}]{shebl2021reappraisal}
\bibinfo{author}{Shebl, A.},  \& \bibinfo{author}{Árpád Csámer}
  (\bibinfo{year}{2021}).
\newblock \bibinfo{title}{Reappraisal of dems, radar and optical datasets in
  lineaments extraction with emphasis on the spatial context}.
\newblock {\it \bibinfo{journal}{Remote Sensing Applications: Society and
  Environment}\/},  {\it \bibinfo{volume}{24}\/}, \bibinfo{pages}{100617}.
  \DOIprefix\doi{10.1016/j.rsase.2021.100617}.
%Type = Article

\bibitem[{Shi et~al.(2021)Shi, Wei, Wei, Wang, Liu \&
  Liu}]{shi2021quantitative}
\bibinfo{author}{Shi, C.}, \bibinfo{author}{Wei, B.}, \bibinfo{author}{Wei, S.}
  et~al. (\bibinfo{year}{2021}).
\newblock \bibinfo{title}{A quantitative discriminant method of elbow point for
  the optimal number of clusters in clustering algorithm}.
\newblock {\it \bibinfo{journal}{EURASIP Journal on Wireless Communications and
  Networking}\/},  {\it \bibinfo{volume}{2021}\/}\bibinfo{issue}{(1)},
  \bibinfo{pages}{1--16}.
%Type = Article

\bibitem[{Shirmard et~al.(2022{\natexlab{a}})Shirmard, Farahbakhsh, Heidari,
  Beiranvand~Pour, Pradhan, M{\"u}ller \& Chandra}]{shirmard2022comparative}
\bibinfo{author}{Shirmard, H.}, \bibinfo{author}{Farahbakhsh, E.},
  \bibinfo{author}{Heidari, E.} et~al. (\bibinfo{year}{2022}{\natexlab{a}}).
\newblock \bibinfo{title}{A comparative study of convolutional neural networks
  and conventional machine learning models for lithological mapping using
  remote sensing data}.
\newblock {\it \bibinfo{journal}{Remote Sensing}\/},  {\it
  \bibinfo{volume}{14}\/}\bibinfo{issue}{(4)}.
%Type = Article

\bibitem[{Shirmard et~al.(2022{\natexlab{b}})Shirmard, Farahbakhsh, M{\"u}ller
  \& Chandra}]{shirmard2022review}
\bibinfo{author}{Shirmard, H.}, \bibinfo{author}{Farahbakhsh, E.},
  \bibinfo{author}{M{\"u}ller, R.~D.} et~al.
  (\bibinfo{year}{2022}{\natexlab{b}}).
\newblock \bibinfo{title}{A review of machine learning in processing remote
  sensing data for mineral exploration}.
\newblock {\it \bibinfo{journal}{Remote Sensing of Environment}\/},  {\it
  \bibinfo{volume}{268}\/}, \bibinfo{pages}{112750}.
%Type = Article

\bibitem[{Sun \& Du(2019)}]{sun2019hyperspectral}
\bibinfo{author}{Sun, W.},  \& \bibinfo{author}{Du, Q.} (\bibinfo{year}{2019}).
\newblock \bibinfo{title}{Hyperspectral band selection: A review}.
\newblock {\it \bibinfo{journal}{IEEE Geoscience and Remote Sensing
  Magazine}\/},  {\it \bibinfo{volume}{7}\/}\bibinfo{issue}{(2)},
  \bibinfo{pages}{118--139}.
%Type = Article

\bibitem[{Tabassum et~al.(2023)Tabassum, Basak, Shao, Haque, Chowdhury \&
  Dey}]{tabassum2023exploring}
\bibinfo{author}{Tabassum, A.}, \bibinfo{author}{Basak, R.},
  \bibinfo{author}{Shao, W.} et~al. (\bibinfo{year}{2023}).
\newblock \bibinfo{title}{Exploring the relationship between land use land
  cover and land surface temperature: A case study in bangladesh and the policy
  implications for the global south}.
\newblock {\it \bibinfo{journal}{Journal of Geovisualization and Spatial
  Analysis}\/},  {\it \bibinfo{volume}{7}\/}\bibinfo{issue}{(2)},
  \bibinfo{pages}{25}. \DOIprefix\doi{10.1007/s41651-023-00155-z}.
%Type = Inproceedings

\bibitem[{Taunk et~al.(2019)Taunk, De, Verma \& Swetapadma}]{taunk2019brief}
\bibinfo{author}{Taunk, K.}, \bibinfo{author}{De, S.}, \bibinfo{author}{Verma,
  S.} et~al. (\bibinfo{year}{2019}).
\newblock \bibinfo{title}{A brief review of nearest neighbor algorithm for
  learning and classification}.
\newblock In {\it \bibinfo{booktitle}{International Conference on Intelligent
  Computing and Control Systems}\/} (pp. \bibinfo{pages}{1255--1260}).
\newblock \DOIprefix\doi{10.1109/ICCS45141.2019.9065747}.
%Type = Inproceedings

\bibitem[{Vincent et~al.(2008)Vincent, Larochelle, Bengio \&
  Manzagol}]{vincent2008extracting}
\bibinfo{author}{Vincent, P.}, \bibinfo{author}{Larochelle, H.},
  \bibinfo{author}{Bengio, Y.} et~al. (\bibinfo{year}{2008}).
\newblock \bibinfo{title}{Extracting and composing robust features with
  denoising autoencoders}.
\newblock In {\it \bibinfo{booktitle}{Proceedings of the 25th international
  conference on Machine learning}\/} (pp. \bibinfo{pages}{1096--1103}).
%Type = Article

\bibitem[{Vincent et~al.(2010)Vincent, Larochelle, Lajoie, Bengio, Manzagol \&
  Bottou}]{vincent2010stacked}
\bibinfo{author}{Vincent, P.}, \bibinfo{author}{Larochelle, H.},
  \bibinfo{author}{Lajoie, I.} et~al. (\bibinfo{year}{2010}).
\newblock \bibinfo{title}{Stacked denoising autoencoders: Learning useful
  representations in a deep network with a local denoising criterion.}
\newblock {\it \bibinfo{journal}{Journal of machine learning research}\/},
  {\it \bibinfo{volume}{11}\/}\bibinfo{issue}{(12)}.
%Type = Article

\bibitem[{Wang et~al.(2016)Wang, Yao \& Zhao}]{wang2016auto}
\bibinfo{author}{Wang, Y.}, \bibinfo{author}{Yao, H.},  \&
  \bibinfo{author}{Zhao, S.} (\bibinfo{year}{2016}).
\newblock \bibinfo{title}{Auto-encoder based dimensionality reduction}.
\newblock {\it \bibinfo{journal}{Neurocomputing}\/},  {\it
  \bibinfo{volume}{184}\/}, \bibinfo{pages}{232--242}.
%Type = Article

\bibitem[{Wang et~al.(2021)Wang, Zuo \& Liu}]{wang2021lithological}
\bibinfo{author}{Wang, Z.}, \bibinfo{author}{Zuo, R.},  \&
  \bibinfo{author}{Liu, H.} (\bibinfo{year}{2021}).
\newblock \bibinfo{title}{Lithological mapping based on fully convolutional
  network and multi-source geological data}.
\newblock {\it \bibinfo{journal}{Remote Sensing}\/},  {\it
  \bibinfo{volume}{13}\/}\bibinfo{issue}{(23)}, \bibinfo{pages}{4860}.
%Type = Inproceedings

\bibitem[{Weilin et~al.(2016)Weilin, Yan \& Shengwei}]{weilin2016application}
\bibinfo{author}{Weilin, Y.}, \bibinfo{author}{Yan, M.},  \&
  \bibinfo{author}{Shengwei, L.} (\bibinfo{year}{2016}).
\newblock \bibinfo{title}{Application of radar and optical remote sensing data
  in lithologic classification and identification}.
\newblock In {\it \bibinfo{booktitle}{IEEE International Geoscience and Remote
  Sensing Symposium}\/} (pp. \bibinfo{pages}{6370--6373}).
%Type = Article

\bibitem[{Wold et~al.(1987)Wold, Esbensen \& Geladi}]{wold1987principal}
\bibinfo{author}{Wold, S.}, \bibinfo{author}{Esbensen, K.},  \&
  \bibinfo{author}{Geladi, P.} (\bibinfo{year}{1987}).
\newblock \bibinfo{title}{Principal component analysis}.
\newblock {\it \bibinfo{journal}{Chemometrics and Intelligent Laboratory
  Systems}\/},  {\it \bibinfo{volume}{2}\/}\bibinfo{issue}{(1-3)},
  \bibinfo{pages}{37--52}.
%Type = Inproceedings

\bibitem[{Xie et~al.(2016)Xie, Girshick \& Farhadi}]{xie2016unsupervised}
\bibinfo{author}{Xie, J.}, \bibinfo{author}{Girshick, R.},  \&
  \bibinfo{author}{Farhadi, A.} (\bibinfo{year}{2016}).
\newblock \bibinfo{title}{Unsupervised deep embedding for clustering analysis}.
\newblock In {\it \bibinfo{booktitle}{International Conference on Machine
  Learning}\/} (pp. \bibinfo{pages}{478--487}).
%Type = Article

\bibitem[{Xu et~al.(2019)Xu, Liang, Zhu, Zheng \& Sun}]{xu2019review}
\bibinfo{author}{Xu, X.}, \bibinfo{author}{Liang, T.}, \bibinfo{author}{Zhu,
  J.} et~al. (\bibinfo{year}{2019}).
\newblock \bibinfo{title}{Review of classical dimensionality reduction and
  sample selection methods for large-scale data processing}.
\newblock {\it \bibinfo{journal}{Neurocomputing}\/},  {\it
  \bibinfo{volume}{328}\/}, \bibinfo{pages}{5--15}.
%Type = Inproceedings

\bibitem[{Yadav et~al.(2019)Yadav, Candela \& Wettergreen}]{yadav2019study}
\bibinfo{author}{Yadav, H.}, \bibinfo{author}{Candela, A.},  \&
  \bibinfo{author}{Wettergreen, D.} (\bibinfo{year}{2019}).
\newblock \bibinfo{title}{A study of unsupervised classification techniques for
  hyperspectral datasets}.
\newblock In {\it \bibinfo{booktitle}{IEEE International Geoscience and Remote
  Sensing Symposium}\/} (pp. \bibinfo{pages}{2993--2996}).
%Type = Article

\bibitem[{Young(2009)}]{young2009ordovician}
\bibinfo{author}{Young, G.~C.} (\bibinfo{year}{2009}).
\newblock \bibinfo{title}{An ordovician vertebrate from western new south
  wales, with comments on cambro-ordovician vertebrate distribution patterns}.
\newblock {\it \bibinfo{journal}{Alcheringa}\/},  {\it
  \bibinfo{volume}{33}\/}\bibinfo{issue}{(1)}, \bibinfo{pages}{79--89}.
%Type = Article

\bibitem[{Yu et~al.(2012)Yu, Porwal, Holden \& Dentith}]{yu2012towards}
\bibinfo{author}{Yu, L.}, \bibinfo{author}{Porwal, A.},
  \bibinfo{author}{Holden, E.-J.} et~al. (\bibinfo{year}{2012}).
\newblock \bibinfo{title}{Towards automatic lithological classification from
  remote sensing data using support vector machines}.
\newblock {\it \bibinfo{journal}{Computers \& Geosciences}\/},  {\it
  \bibinfo{volume}{45}\/}, \bibinfo{pages}{229--239}.
%Type = Article

\bibitem[{Yuan \& Yang(2019)}]{yuan2019research}
\bibinfo{author}{Yuan, C.},  \& \bibinfo{author}{Yang, H.}
  (\bibinfo{year}{2019}).
\newblock \bibinfo{title}{Research on k-value selection method of k-means
  clustering algorithm}.
\newblock {\it \bibinfo{journal}{J}\/},  {\it
  \bibinfo{volume}{2}\/}\bibinfo{issue}{(2)}, \bibinfo{pages}{226--235}.
%Type = Article

\bibitem[{Zhang et~al.(2016{\natexlab{a}})Zhang, Zhai, Zhang \&
  Li}]{zhang2016spectral}
\bibinfo{author}{Zhang, H.}, \bibinfo{author}{Zhai, H.},
  \bibinfo{author}{Zhang, L.} et~al. (\bibinfo{year}{2016}{\natexlab{a}}).
\newblock \bibinfo{title}{Spectral--spatial sparse subspace clustering for
  hyperspectral remote sensing images}.
\newblock {\it \bibinfo{journal}{IEEE Transactions on Geoscience and Remote
  Sensing}\/},  {\it \bibinfo{volume}{54}\/}\bibinfo{issue}{(6)},
  \bibinfo{pages}{3672--3684}.
%Type = Article

\bibitem[{Zhang et~al.(2018{\natexlab{a}})Zhang, Yu \& Tao}]{zhang2018local}
\bibinfo{author}{Zhang, J.}, \bibinfo{author}{Yu, J.},  \&
  \bibinfo{author}{Tao, D.} (\bibinfo{year}{2018}{\natexlab{a}}).
\newblock \bibinfo{title}{Local deep-feature alignment for unsupervised
  dimension reduction}.
\newblock {\it \bibinfo{journal}{IEEE Transactions on Image Processing}\/},
  {\it \bibinfo{volume}{27}\/}\bibinfo{issue}{(5)},
  \bibinfo{pages}{2420--2432}. \DOIprefix\doi{10.1109/TIP.2018.2804218}.
%Type = Article

\bibitem[{Zhang et~al.(2016{\natexlab{b}})Zhang, Yi, Li, Wang, Tang, Zhong, Li,
  Wang \& Bie}]{zhang2016integrating}
\bibinfo{author}{Zhang, T.}, \bibinfo{author}{Yi, G.}, \bibinfo{author}{Li, H.}
  et~al. (\bibinfo{year}{2016}{\natexlab{b}}).
\newblock \bibinfo{title}{Integrating data of aster and landsat-8 oli (ao) for
  hydrothermal alteration mineral mapping in duolong porphyry cu-au deposit,
  tibetan plateau, china}.
\newblock {\it \bibinfo{journal}{Remote Sensing}\/},  {\it
  \bibinfo{volume}{8}\/}\bibinfo{issue}{(11)}, \bibinfo{pages}{890}.
%Type = Article

\bibitem[{Zhang et~al.(2018{\natexlab{b}})Zhang, Jiang, Li \&
  Yang}]{zhang2018automated}
\bibinfo{author}{Zhang, Z.}, \bibinfo{author}{Jiang, T.}, \bibinfo{author}{Li,
  S.} et~al. (\bibinfo{year}{2018}{\natexlab{b}}).
\newblock \bibinfo{title}{Automated feature learning for nonlinear process
  monitoring--an approach using stacked denoising autoencoder and k-nearest
  neighbor rule}.
\newblock {\it \bibinfo{journal}{Journal of Process Control}\/},  {\it
  \bibinfo{volume}{64}\/}, \bibinfo{pages}{49--61}.
%Type = Article

\bibitem[{Zhao et~al.(2023)Zhao, Zhang, Tang, Luo, Wan \&
  An}]{zhao2023recognition}
\bibinfo{author}{Zhao, B.}, \bibinfo{author}{Zhang, D.}, \bibinfo{author}{Tang,
  P.} et~al. (\bibinfo{year}{2023}).
\newblock \bibinfo{title}{Recognition of multivariate geochemical anomalies
  using a geologically-constrained variational autoencoder network with
  spectrum separable module--a case study in shangluo district, china}.
\newblock {\it \bibinfo{journal}{Applied Geochemistry}\/},  {\it
  \bibinfo{volume}{156}\/}, \bibinfo{pages}{105765}.
%Type = Article

\bibitem[{Zhong et~al.(2021)Zhong, Wang, Wang \& Zhang}]{zhong2021advances}
\bibinfo{author}{Zhong, Y.}, \bibinfo{author}{Wang, X.}, \bibinfo{author}{Wang,
  S.} et~al. (\bibinfo{year}{2021}).
\newblock \bibinfo{title}{Advances in spaceborne hyperspectral remote sensing
  in china}.
\newblock {\it \bibinfo{journal}{Geo-spatial Information Science}\/},  {\it
  \bibinfo{volume}{24}\/}\bibinfo{issue}{(1)}, \bibinfo{pages}{95--120}.
%Type = Article

\bibitem[{Zhou et~al.(2019)Zhou, Han, Cheng \& Zhang}]{zhou2019learning}
\bibinfo{author}{Zhou, P.}, \bibinfo{author}{Han, J.}, \bibinfo{author}{Cheng,
  G.} et~al. (\bibinfo{year}{2019}).
\newblock \bibinfo{title}{Learning compact and discriminative stacked
  autoencoder for hyperspectral image classification}.
\newblock {\it \bibinfo{journal}{IEEE Transactions on Geoscience and Remote
  Sensing}\/},  {\it \bibinfo{volume}{57}\/}\bibinfo{issue}{(7)},
  \bibinfo{pages}{4823--4833}.
%Type = Article

\bibitem[{Zuo et~al.(2019)Zuo, Xiong, Wang \& Carranza}]{zuo2019deep}
\bibinfo{author}{Zuo, R.}, \bibinfo{author}{Xiong, Y.}, \bibinfo{author}{Wang,
  J.} et~al. (\bibinfo{year}{2019}).
\newblock \bibinfo{title}{Deep learning and its application in geochemical
  mapping}.
\newblock {\it \bibinfo{journal}{Earth-science reviews}\/},  {\it
  \bibinfo{volume}{192}\/}, \bibinfo{pages}{1--14}.

\end{thebibliography}

\end{document}